\documentclass[11pt]{article}
\usepackage{enumitem}
\usepackage{booktabs}
\usepackage{multirow}
\usepackage{multicol}
\usepackage{amsmath}
\usepackage{mathtools}
\usepackage{subcaption}
\usepackage{array}
\usepackage{hyperref}
\usepackage[table]{xcolor}
\usepackage{amsfonts}
\usepackage{placeins}
\usepackage[most]{tcolorbox}
\usepackage{wrapfig}
\usepackage{pifont}
\usepackage{capt-of} 
\usepackage{caption}
\usepackage{siunitx}
\usepackage{marvosym}  % Package for symbols
\usepackage{stfloats} % add this to your preamble
\usepackage{cuted}
\usepackage{tabularx}
\usepackage{titletoc}  % 如果真有用就保留
\usepackage{etoc}      % 放后面

\usepackage{acl}

\usepackage{times}

\usepackage{latexsym}

\usepackage[T1]{fontenc}

\usepackage[utf8]{inputenc}

\usepackage{microtype}

\usepackage{inconsolata}

\usepackage{graphicx}

\title{Not Just Reason, Not Just Scan:
    Reinforcement Learning for Proactive Scientific Error Verification over Academic Papers}

\author{
  Rongjin Li$^{1}$,
  Yuanxin Liu$^{2}$,
  Hao Zhou$^{3}$,
  Fandong Meng$^{3}$,
  Jie Zhou$^{3}$,
  Xu Sun$^{2}$\thanks{Corresponding author.} \\[2mm]
  $^{1}$Beijing University of Posts and Telecommunications \\
  $^{2}$State Key Laboratory for Multimedia Information Processing, \\
  School of Computer Science, Peking University \\
  $^{3}$WeChat AI, Tencent Inc., China \\
  \small\texttt{lirongjin@bupt.edu.cn},
  \texttt{xusun@pku.edu.cn}
}

\begin{document}
\maketitle
\begin{abstract}
Multimodal large language models (MLLMs) are increasingly capable scientific assistants, yet they remain far from fully autonomous research. This transition requires models to actively inspect academic papers, build global evidence views, and make traceable judgments without prespecified issues or evidence. However, existing work provides limited task paradigms or training studies for such issue- and evidence-absent verification. We study this challenge through scientific error detection, where models must determine whether errors exist and justify them with evidence-based reasoning. To fill this gap, we present VERA-RL, a reinforcement-learning formulation for scientific error detection over academic papers. Following a \textit{Reason}--\textit{Verify}--\textit{Scan} progression, we construct VERA-13K, a 12,900-sample dataset organized into 4,300 matched chains, covering 6 scientific-error categories across the research workflow and broad natural-science domains. We further introduce fine-grained rewards for reasoning completeness, evidence alignment, and error precision. Training Qwen3-VL-8B with VERA-RL substantially improves verifiable reasoning, approaching flagship MLLMs such as Gemini 3 Pro and Qwen3-VL-235B-A22B on \textit{Scan}.
\end{abstract}

\begin{figure}[t]
  \centering
  \vspace{-0.5em}
  \includegraphics[width=0.95\linewidth]{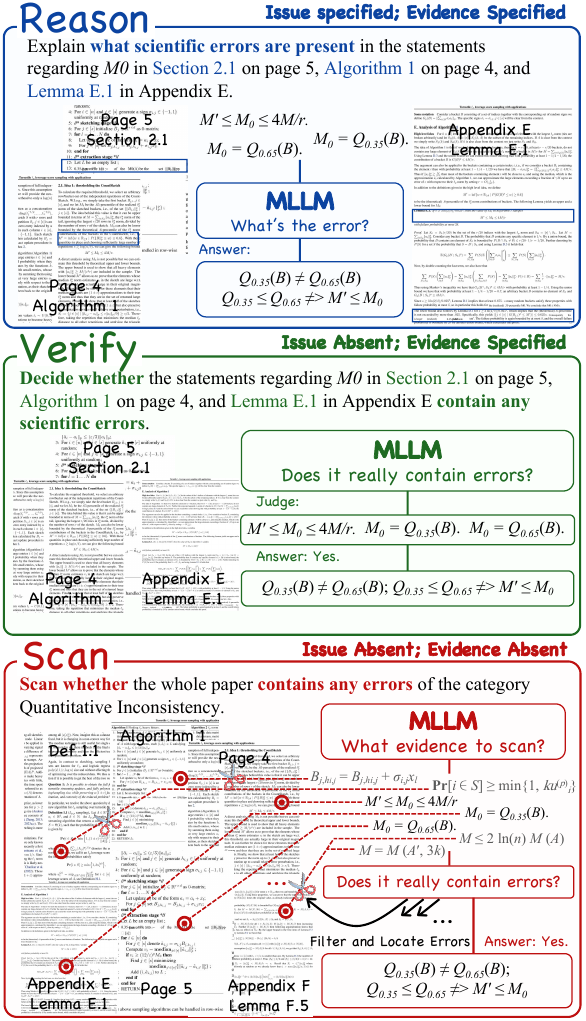}
  \vspace{-0.7em}
  \caption{
    Overview of \textit{Reason}-\textit{Verify}-\textit{Scan} paradigm.
  }
  \label{fig:Figure1}
  \vspace{-0.6em}
\end{figure}
\section{Introduction}
End-to-end independent scientific research by multimodal large language models (MLLMs)~\cite{bai2025qwen3vltechnicalreport,bytedance2025seed16tech,gpt54,anthropic2025claude4} is a long-standing goal of Deep Research and a milestone toward AGI~\cite{ge2023openagi,morris2024position}. 
In recent years, MLLMs have become increasingly embedded in the research workflow, automating various aspects of scientific work~\cite{he2025pasallmagentcomprehensive}. 
For example, Google Deep Research~\cite{comanici2025gemini25pushingfrontier} aggregates multi-source materials into structured reports, while Paper Decision~\cite{PaperDecision2026} synthesizes reviews and rebuttals with multi-agent systems to predict final decisions. 
Together, these systems point to a broader shift from passive information synthesis toward active scientific inquiry.

As research tasks move beyond answering given questions or summarizing existing work, paper-level judgment becomes central to autonomous research: deciding what should be checked, gathering dispersed evidence, and assessing whether claims are supported by the paper.
Such judgment underlies research-quality assessment, idea discovery, and refinement of scientific work. 

However, current MLLMs remain far from this goal. 
A key limitation is that most existing methods still operate in specified settings, where both the target problem and the supporting evidence are provided in advance~\cite{xu-etal-2025-llms-identify,wang-etal-2025-sciver}. 
Recently, ScholScan~\cite{li2026not} explicitly defined the \textit{Scan} task paradigm, which requires models to perform global verification and consistency checking across the full text, rather than relying on evidence anchored in specific sections. However, existing work still provides limited discussion on methodologies and training paths for systematically enhancing the \textit{Scan} capability.

%Building on ScholScan, we study how to systematically train MLLMs for \textit{Scan} rather than treating it only as an evaluation task. We use scientific error detection as the concrete task, since it naturally requires models to conduct scanning verification and consistency checking without target cues. To this end, we propose \textsc{VERA-RL}, a training formulation that couples progressive task decomposition with reinforcement learning (RL) rewards for verifiable reasoning over academic papers~\cite{rafailov2024directpreferenceoptimizationlanguage,schulman2017proximalpolicyoptimizationalgorithms,shao2024deepseekmathpushinglimitsmathematical,yu2025dapoopensourcellmreinforcement}.

%In our experiments, we observe that the capabilities learned by existing models in traditional reasoning tasks with specified evidence do not naturally transfer to \textit{Scan} tasks, leading to a significant performance gap. This makes direct end-to-end training on the full \textit{Scan} setting less effective. To bridge this gap, we decompose \textit{Scan} into three progressive capability stages: \textit{Reason}, \textit{Verify}, and \textit{Scan} (Figure \ref{fig:Figure1}). Together, these stages provide a path from evidence-conditioned reasoning to global scanning verification.

Building on ScholScan, we study how to systematically train MLLMs for \textit{Scan} rather than treating it only as an evaluation task. We instantiate this setting as scientific error detection and propose \textsc{VERA-RL}, a training formulation that couples progressive task decomposition with reinforcement learning (RL) rewards for verifiable reasoning over academic papers~\cite{rafailov2024directpreferenceoptimizationlanguage,schulman2017proximalpolicyoptimizationalgorithms,shao2024deepseekmathpushinglimitsmathematical,yu2025dapoopensourcellmreinforcement}. To support training, we construct \textsc{VERA-13K} from controlled edits on accepted papers and objective errors extracted from peer reviews, converting each error into \textit{Reason}--\textit{Verify}--\textit{Scan} chains (Figure~\ref{fig:Figure1}) across six categories.

In our experiments, we observe that evidence-specified reasoning does not naturally transfer to \textit{Scan}, leading to a cue-removal gap and making direct end-to-end training on the full \textit{Scan} setting less effective. \textsc{VERA-RL} substantially improves Qwen3-VL-8B on \textsc{VERA-13K} and shows measurable transfer to ScholScan. Further analyses show that \textit{Reason} provides an internal-knowledge reference, while ablations confirm that both staged tasks and multi-dimensional rewards are necessary for stable gains.

In summary, our contributions are as follows:

\begin{itemize}[leftmargin=*,itemsep=2pt,topsep=2pt]
  \item We formulate a three-stage \textit{Reason}--\textit{Verify}--\textit{Scan} paradigm that turns \textit{Scan} into a trainable progression from evidence-specified reasoning to issue- and evidence-absent verification.
  \item We build \textsc{VERA-13K} with a reusable construction pipeline and 12,900 filtered samples covering 6 scientific-error categories across typical risk points in the research process and a wide range of natural-science domains.
  \item We introduce fine-grained rewards for reasoning completeness, evidence alignment, and error precision, enabling RL to target the core dimensions required by scientific error verification.
  \item Experiments show that \textsc{VERA-RL} substantially improves Qwen3-VL-8B-Instruct and reaches performance comparable to the Qwen3-VL-235B-A22B series. Ablations further show that both the staged task paradigm and the reward system are essential for stable gains.
\end{itemize}

\section{Related Work}
\subsection{Academic Paper Understanding}
Academic papers differ from general documents in their dense domain knowledge and rigorous logic. While previous work focused on isolated elements such as paragraphs or figures, it often overlooked the added challenge of reasoning over complete documents.~\cite{chen2026rpcbench,wang2024charxivchartinggapsrealistic,Auer2023TheSS,li-etal-2024-multimodal-arxiv} Recent studies have adopted full-document inputs, but often treat the paper as a sparse mix of key passages and irrelevant text.~\cite{ma2024mmlongbenchdoc,yan2025mmcradvancingvisuallanguage,lou2025aaar,zhao-etal-2024-docmath} This framing narrows the evaluation scope, reducing it to long-context retrieval paired with localized reasoning.

More critically, most benchmarks still adopt a QA paradigm, diverging from real-world scientific tasks. Efforts like PRISMM-Bench~\cite{selch2025prismmbenchbenchmarkpeerreviewgrounded,xi2025flawsbenchmarkerroridentification,tu2026paperauditbenchbenchmarkingerrordetection} simulate reviewer-style understanding, yet still embed explicit clues and presuppose answer existence. ScholScan~\cite{li2026not} introduces the \textit{Scan}-oriented task, targeting assumption-absent and evidence-absent conditions. However, it treats \textit{Scan} as a static capability without examining how it can be effectively trained. A summary table comparing representative benchmarks is provided in Appendix~\ref{app:appendix_a}.

\subsection{RL for LLM Reasoning}
Since DeepSeek-R1 and OpenAI-o1~\cite{guo2025deepseek,openai2024openaio1card}, large-scale reinforcement learning has become a key approach for improving the reasoning capabilities of LLMs. Recent studies have extended RL to more complex settings, including multimodal and long-context inputs. However, their task and data designs are not yet well aligned with document-level scientific verification. LoongRL and QwenLong-L1~\cite{wang2025loongrlreinforcementlearningadvanced,wan2025qwenlongl1longcontextlargereasoning} extend RL to long-context settings through paragraph concatenation, while VRAG-RL~\cite{wang2025vragrlempowervisionperceptionbasedrag} incorporates retrieval into the RL pipeline. These designs improve long-context grounding or retrieval-conditioned reasoning, but provide limited guidance for training \textit{Scan}-style verification over full academic papers.

\section{Methodology}
\begin{figure*}[t]
  \centering
  \includegraphics[width=\linewidth]{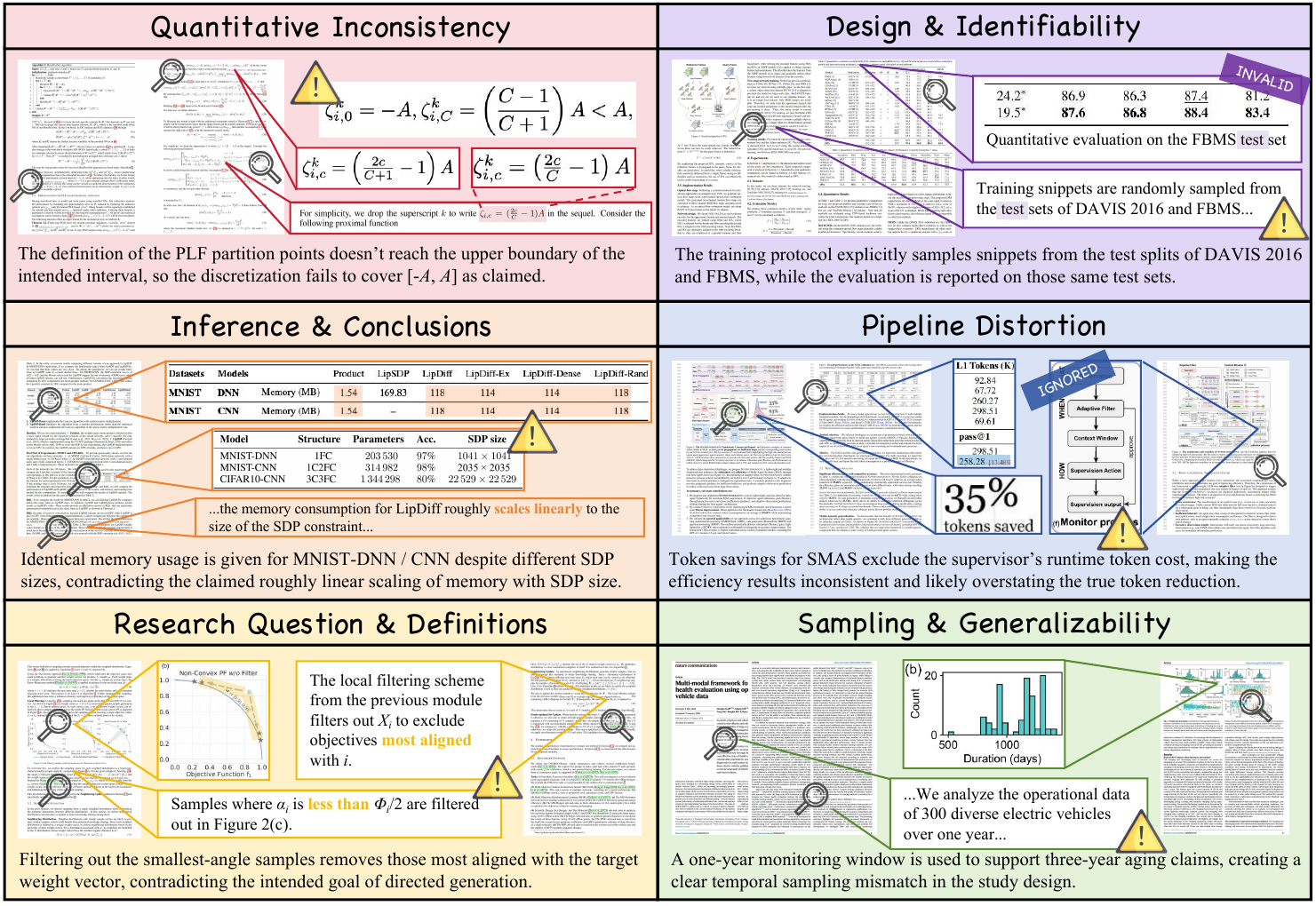}
  \caption{
    Sampled \textsc{VERA-13K} examples with 6 error types. We consolidate and extend the taxonomy proposed in ScholScan, resulting in a more coherent and well-structured categorization scheme.
  }
  \label{fig:Figure2}
\end{figure*}
\begin{table*}[h]
\caption{Detailed statistics and distributions of our train and test datasets. Length is calculated by the Qwen3 tokenizer. \textit{Scan}, \textit{Verify}, and \textit{Reason} are strictly balanced in every split and are constructed as a chained curriculum from the same underlying instances. \textbf{QI}: Quantitative Inconsistency; \textbf{DI}: Design \& Identifiability; \textbf{IC}: Inference \& Conclusions; \textbf{PD}: Pipeline Distortion; \textbf{RQD}: Research Question \& Definitions; \textbf{SG}: Sampling \& Generalizability.}
\label{tab:split}
\centering
\setlength{\tabcolsep}{6pt}
\renewcommand{\arraystretch}{1}

\small

\begin{tabular}{l|cc|ccccccc}
\toprule
\multirow{2}{*}{\textbf{Statistics}} &
\multicolumn{2}{c|}{\textbf{Train Dataset}} &
\multicolumn{7}{c}{\textbf{Test Dataset}} \\
& \textbf{SFT} & \textbf{RL} & \textbf{Avg.}
 & \textbf{QI} & \textbf{DI} & \textbf{IC} & \textbf{PD} & \textbf{RQD} & \textbf{SG} \\
\midrule
\# Examples   & 10,500 & 1,500 & - & 150 & 150 & 150 & 150 & 150 & 150 \\
Avg.\ Length  & 31,134 & 30,649 & 31,837 & 29,956 & 37,066 & 31,985 & 31,100 & 32,792 & 28,124 \\
Avg.\ Evidence  & 3.54 & 3.52 & 3.58 & 3.28 & 3.28 & 3.80 & 3.60 & 3.92 & 3.62 \\
Avg.\ Reasoning Step  & 3.81 & 3.79 & 3.79 & 3.77 & 3.85 & 3.72 & 3.84 & 3.81 & 3.74 \\
\bottomrule

\end{tabular}
\end{table*}

\subsection{Task Definition}
We formulate scientific error detection as producing a structured answer from a paper input, consisting of evidence points, reasoning steps, and a final judgment. Based on the availability of issue and evidence cues, we define 3 stages: \textit{Reason} assumes an error with specified evidence; \textit{Verify} provides candidate evidence without assuming an error; and \textit{Scan} is both issue- and evidence-absent, specifying only a broad scanning target such as an error type. Question examples are shown in Figure~\ref{fig:Figure1}, and the prompt and answer format is provided in Table~\ref{qa_format}.

Each underlying error instance can be converted into a matched \textit{Reason}--\textit{Verify}--\textit{Scan} chain by rewriting the query while preserving the corresponding evidence and reasoning structure.
\begin{table}[h]
\caption{QA format used in \textsc{VERA-13K}.}
\label{qa_format}
\centering
\small
\setlength{\tabcolsep}{0pt}
\begin{tabularx}{\columnwidth}{@{}X@{}}
\toprule

\textbf{Question Prompt:}\par
\quad Read the provided paper and answer the question. The question will ask you to check and reason about scientific errors in specific parts of the paper. In your answer, show your reasoning process and explain in detail the exact nature of the errors.\par
\quad \texttt{\textcolor{blue}{<Question> }} \texttt{\textcolor{red}{<Interleaved context of the paper> }}

\\
\midrule

\textbf{Gold Answer:}\par
Evidence: (Only for \textit{Scan} questions)\par
\texttt{- <Evidence 1>} \quad \quad \texttt{- ...} \quad \quad \texttt{- <Evidence \textit{n}>}

Reasoning:\par
\texttt{- <Step 1>} \quad \quad \quad \quad \texttt{- ...} \quad \quad \quad \texttt{- <Step \textit{n}>}

Answer: \texttt{...}

\\
\bottomrule
\end{tabularx}
\end{table}
\vspace{-1.0em}

\subsection{Dataset Construction}

\paragraph{Overview} Following ScholScan, we consolidate and extend its error taxonomy into 6 categories aligned with major failure points in the research workflow (Figure~\ref{fig:Figure2}). We construct \textsc{VERA-13K}, a dataset of 12,900 samples for training and evaluation, derived by rewriting 4,300 errors into \textit{Reason}--\textit{Verify}--\textit{Scan} chains. Table~\ref{tab:split} summarizes the splits and composition of \textsc{VERA-13K}, where each subset consists of complete three-stage task chains.

\paragraph{Data Collection}
We collect data from two complementary sources. First, we curate accepted papers from top-tier journals (\textit{e.g.}, Nature Communications) and premier conferences (\textit{e.g.}, ICML), which provide high-quality and clean materials. Second, we incorporate reviews from ICLR submissions, which provide human-written feedback on potential weaknesses in submitted papers. Additional statistics are provided in Appendix \ref{app:appendix_a}.

\paragraph{Data Generation and Quality Control} For accepted papers, we instruct Gemini 3 Flash~\cite{gemini3} to inject scientific errors through paragraph-level edits spanning multiple sections, under strict category definitions and constraints. For review-derived papers, it extracts objective scientific errors while filtering out subjective feedback. Each instance is rewritten into a \textit{Reason}--\textit{Verify}--\textit{Scan} chain, with formulations standardized as in Table~\ref{qa_format}. We then use Seed-1.6-Thinking for Pass@4 filtering, retaining a sample only judged as correct or partially correct. This step filters out unverifiable or weakly supported annotations. Prompts and other details are provided in our repository; see Appendix \ref{app:appendix_h}.

\subsection{RL for Verifiable Reasoning}
\paragraph{Algorithm}
We optimize the policy with DAPO, a mature RL algorithm suitable for the structured outputs required by academic-paper verification, and use it to study whether the \textit{Scan} capability can be learned through \textit{Reason}--\textit{Verify}--\textit{Scan} staged tasks.  For each training sample $(q,L,a)$, where $q$ is the question, $L$ is the input paper, and $a$ is the gold answer, DAPO samples $G$ trajectories $\{y_i\}_{i=1}^{G}$ from the old policy $\pi_{\theta_{\text{old}}}$ and updates the current policy $\pi_\theta$ by maximizing:

\begin{equation*}
\small
\begin{aligned}
J_{\text{DAPO}}(\theta) =\; & \mathbb{E}_{(q, L) \sim \mathcal{D},\, \{y_i\}_{i=1}^G \sim \pi_{\theta_{\text{old}}}} 
\left[ \frac{1}{\sum_{i=1}^{G} |y_i|} \right] \cdot \Bigg( \\
& \sum_{i=1}^{G} \sum_{t=1}^{|y_i|} 
\min \Bigg(
\frac{\pi_{\theta}(y_{i,t} \mid q, L, y_{i,<t})}
     {\pi_{\theta_{\text{old}}}(y_{i,t} \mid q, L, y_{i,<t})} A_{i,t}, \\
& \hspace{-48pt}
\text{clip}\left(
\frac{\pi_{\theta}(y_{i,t} \mid q, L, y_{i,<t})}
     {\pi_{\theta_{\text{old}}}(y_{i,t} \mid q, L, y_{i,<t})},\, 1 - \epsilon_{\text{low}},\, 1 + \epsilon_{\text{high}}
\right) A_{i,t}
\Bigg) \Bigg) \\
& - \beta \cdot \mathrm{D_{KL}}\left[
\pi_{\theta}(\cdot \mid q, L) \,\|\, \pi_{\text{ref}}(\cdot \mid q, L)
\right]
\end{aligned}
\end{equation*}

Here $A_{i,t}$ is the token-level advantage for trajectory $y_i$ at token $t$, $\epsilon_{\mathrm{low}}$ and $\epsilon_{\mathrm{high}}$ are asymmetric clipping bounds, $\beta$ controls the KL penalty, and $\pi_{\mathrm{ref}}$ is the reference policy. The advantage is normalized within the sampled trajectory group:

\begin{equation*}
\small
A_{i,t} = \frac{r_i - \mathrm{mean}(\{r_i\}_{i=1}^{G})}{\mathrm{std}(\{r_i\}_{i=1}^{G})}
\end{equation*}

where $r_i$ is the reward assigned to trajectory $y_i$.

\paragraph{Rewards} For a given question $q$, reference answer $a$, and model trajectory $y$, we design rewards to capture both process-level correctness and task-specific verification requirements. $R_{\text{completeness}}$ extends conventional answer correctness to the reasoning process. Since scientific error detection often admits partially correct answers, it measures the proportion of reference answer points covered by the trajectory rather than only judging the final answer. Beyond this correctness-oriented signal, \textit{Scan}-style verification imposes two additional requirements. First, the model must ground its judgment in the paper rather than produce unsupported explanations, so $R_{\text{alignment}}$ measures the overlap between generated and reference evidence points. Second, open-ended error detection may incentivize excessive candidate errors, so $R_{\text{precision}}$ penalizes unsupported error claims. Together, these terms define rewards for reasoning completeness, evidence grounding, and error precision:

\begin{equation*}
\begin{aligned}
R_{\text{final}}
&= \omega_1 R_{\text{completeness}}
 + \omega_2 R_{\text{alignment}}
 + \omega_3 R_{\text{precision}} \\
&= \omega_1
\frac{\lvert \hat{\mathcal{R}} \cap \mathcal{R}^{*} \rvert}
{\lvert \mathcal{R}^{*} \rvert}
+ \omega_2
\frac{2\lvert \hat{\mathcal{E}} \cap \mathcal{E}^{*} \rvert}
{\lvert \hat{\mathcal{E}} \rvert + \lvert \mathcal{E}^{*} \rvert} \\
&\quad + \omega_3 \mathbb{I}_{\mathrm{error}} e^{-0.4m}.
\end{aligned}
\end{equation*}

Here $\hat{\mathcal{R}}$ and $\hat{\mathcal{E}}$ denote the reasoning points and evidence points extracted from the trajectory, while $\mathcal{R}^{*}$ and $\mathcal{E}^{*}$ denote the corresponding reference points. $m$ denotes the number of unsupported error claims, and $\mathbb{I}_{\mathrm{error}}$ is 1 for an error prediction and 0 otherwise. These quantities are extracted and matched by Seed-1.6-Thinking as a fixed structured evaluator, while the final reward is computed by the above rule rather than directly assigned by the evaluator. This structured evaluation is less dependent on unconstrained LLM-as-a-Judge preferences.

For \textit{Reason} and \textit{Verify}, evidence is specified, so we disable $R_{\text{alignment}}$ and set $(\omega_1,\omega_2,\omega_3)=(0.6,0,0.4)$, keeping reasoning as the main target while using precision as a constraint. For \textit{Scan}, where evidence must be identified rather than given, we set $(0.4,0.4,0.2)$ so that $R_{\text{alignment}}$ receives a role comparable to $R_{\text{completeness}}$. Appendix~\ref{app:appendix_e} reports statistics supporting this choice.
\section{Experiments}
\subsection{Setup}
\newcommand{\ScanMetricBlock}[1]{%
  \rowcolor{blue!10}%
  \multicolumn{8}{c}{\textbf{Scan}\quad $#1$}\\[-0.05ex]
}

\begin{table*}[t]
\caption{\textit{Scan}-task evaluation results (scaled by 100) for baselines and models trained in the main experiments.}
\label{tab:main_results}
\centering
{%
\small
\setlength{\tabcolsep}{5pt}
\setlength{\extrarowheight}{0.25ex}
\renewcommand{\arraystretch}{1.08}
\begin{tabular}{>{\raggedright\arraybackslash}p{5cm} *{7}{>{\centering\arraybackslash}p{1cm}}}
\toprule
\textbf{Models} & \textbf{Avg.} & \textbf{QI} & \textbf{DI} & \textbf{IC} & \textbf{PD} & \textbf{RQD} & \textbf{SG} \\
\midrule

\ScanMetricBlock{R_{\mathrm{completeness}}}
Gemini 3 Pro & 22.8 & 24.4 & 13.2 & 30.0 & 17.7 & 25.8 & 26.0 \\
Qwen3-VL-235B-A22B (Thinking) & 16.7 & 22.2 & 12.5 & 21.7 & 10.2 & 13.3 & 20.3 \\
Qwen3-VL-235B-A22B (Instruct) & 3.4 & 7.0 & 3.3 & 7.8 & 1.5 & 0.0 & 0.5 \\
Qwen3-VL-8B (Instruct) & 1.5 & 0.0 & 0.0 & 5.3 & 1.5 & 0.0 & 2.0 \\
\textbf{Qwen3-VL-8B (SFT, ours)} & 5.3 & 5.2 & 1.3 & 5.8 & 6.3 & 3.0 & 10.3 \\
\textbf{Qwen3-VL-8B (SFT+RL, ours)} & 8.2 & 7.8 & 0.5 & 10.0 & 6.8 & 11.3 & 12.8 \\

\ScanMetricBlock{R_{\mathrm{alignment}}}
Gemini 3 Pro & 17.8 & 22.7 & 9.4 & 25.7 & 15.0 & 17.4 & 16.9 \\
Qwen3-VL-235B-A22B (Thinking) & 13.0 & 21.4 & 8.9 & 17.5 & 7.1 & 11.0 & 12.2 \\
Qwen3-VL-235B-A22B (Instruct) & 2.4 & 4.7 & 2.0 & 6.3 & 0.8 & 0.0 & 0.8 \\
Qwen3-VL-8B (Instruct) & 1.0 & 0.0 & 0.0 & 3.9 & 2.0 & 0.0 & 0.0 \\
\textbf{Qwen3-VL-8B (SFT, ours)} & 2.5 & 1.1 & 0.8 & 5.3 & 1.1 & 0.8 & 4.8 \\
\textbf{Qwen3-VL-8B (SFT+RL, ours)} & 6.2 & 9.2 & 1.0 & 7.7 & 4.5 & 6.4 & 8.4 \\

\ScanMetricBlock{R_{\mathrm{precision}}}
Gemini 3 Pro & 40.4 & 45.7 & 43.5 & 41.3 & 42.1 & 35.3 & 34.4 \\
Qwen3-VL-235B-A22B (Thinking) & 27.5 & 38.2 & 32.1 & 26.5 & 26.2 & 21.3 & 20.9 \\
Qwen3-VL-235B-A22B (Instruct) & 5.4 & 6.8 & 3.8 & 19.0 & 0.4 & 0.0 & 2.3 \\
Qwen3-VL-8B (Instruct) & 5.0 & 3.0 & 3.3 & 14.8 & 2.2 & 2.3 & 4.5 \\
\textbf{Qwen3-VL-8B (SFT, ours)} & 56.3 & 52.3 & 57.1 & 55.7 & 54.0 & 58.3 & 60.2 \\
\textbf{Qwen3-VL-8B (SFT+RL, ours)} & 68.6 & 67.0 & 65.2 & 68.9 & 70.4 & 69.6 & 70.3 \\

\ScanMetricBlock{R_{\mathrm{final}}}
GPT-5.4 & 29.9 & 49.9 & 26.2 & 22.8 & 31.8 & 26.2 & 23.8 \\
Gemini 3 Pro & 24.3 & 28.0 & 17.7 & 30.3 & 21.5 & 24.4 & 24.0 \\
Seed-1.6-Thinking & 21.2 & 25.9 & 17.1 & 27.5 & 16.1 & 22.6 & 17.6 \\
Qwen3-VL-Plus & 26.2 & 35.7 & 17.7 & 30.0 & 22.1 & 24.5 & 27.3 \\
Qwen3-VL-235B-A22B (Thinking) & 17.4 & 25.1 & 15.0 & 21.0 & 12.2 & 14.0 & 17.2 \\
Qwen3-VL-235B-A22B (Instruct) & 3.4 & 6.0 & 2.9 & 9.4 & 1.0 & 0.0 & 1.0 \\
%Qwen3-VL-32B & 21.2 & 21.1 & 15.9 & 26.1 & 20.5 & 20.8 & 22.6 \\
Qwen3-VL-8B (Instruct) & 2.0 & 0.6 & 0.7 & 6.7 & 1.8 & 0.5 & 1.7 \\
\textbf{Qwen3-VL-8B (SFT, ours)} & 14.4 & 13.0 & 12.3 & 15.6 & 14.2 & 13.2 & 18.1 \\
\textbf{Qwen3-VL-8B (SFT+RL, ours)} & 19.5 & 20.2 & 13.6 & 20.9 & 18.6 & 21.0 & 22.5 \\

\bottomrule
\end{tabular}
}
\end{table*}
\paragraph{Inputs} Prior work typically relies on either plain-text OCR, which discards all visual elements, or full-page image rendering, which demands high resolution to retain dense textual content.~\cite{ma2024mmlongbenchdoc,selch2025prismmbenchbenchmarkpeerreviewgrounded} These approaches fall short for our task, where figure-text alignment and layout preservation are essential. We instead adopt DeepSeek-OCR~\cite{wei2025deepseekocrcontextsopticalcompression} to convert papers into interleaved text and images, preserving most of the original information. 
%This format maintains a balance between information density and input cost, while naturally aligning with human reading habits.

\paragraph{Training Setup} We train Qwen3-VL-8B-Instruct with 1 epoch of SFT using a batch size of 8, followed by 30 RL steps with a batch size of 32 and random sampling over all task types. Further details are provided in Appendix~\ref{app:appendix_b}.

\paragraph{Baselines}
We evaluate 8 models spanning flagship proprietary models and Qwen3-VL variants. For selected Qwen3-VL models, we compare their Instruct and Thinking versions to assess how training schemes affect verifiable reasoning.

\begin{table*}[t]
\caption{
Evaluation results (scaled by 100) for baselines and models trained in the main experiments.
$S_{\text{reason}}$, $S_{\text{location}}$, $P_{\text{unrelated\_err}}$, and $S(m)$ are evaluation metrics introduced in ScholScan, with detailed definitions provided in Appendix~\ref{app:appendix_c}.
$R_{\text{precision}}$ is not applicable in this evaluation setting and is therefore omitted.
}
\centering
\setlength{\tabcolsep}{6pt}
\renewcommand{\arraystretch}{1}
\small
\begin{tabular}{ccccccc}

\toprule
\textbf{Models} & $R_{\text{completeness}}$ & $R_{\text{alignment}}$ & $S_{\text{reason}}$ & $S_{\text{location}}$ & $P_{\text{unrelated\_err}}$ & $S(m)$\\
\midrule
Gemini 3 Pro                & 11.3 & 7.2 & 11.0 & 6.0 & 10.2 & 5.5\\
Qwen3-VL-235B-A22B-Thinking    & 4.7 & 2.8 & 4.6 & 2.4 & 5.3 & 2.6\\
Qwen3-VL-235B-A22B-Instruct    & 0.8 & 0.4 & 0.8 & 0.3 & 0.4 & 0.1\\
Qwen3-VL-8B-Instruct          & 0.0 & 0.0 & 0.0 & 0.0 & 0.0 & 0.0\\
Qwen3-VL-8B (SFT)             & 0.4 & 0.1 & 0.4 & 0.0 & 3.8 & 0.0\\
Qwen3-VL-8B (SFT+RL) & 0.5 & 0.3 & 0.5 & 0.2 & 0.5 & 0.2 \\
\bottomrule
\label{ScholScan}
\end{tabular}
\end{table*}
\subsection{Main Results}
Table~\ref{tab:main_results} reports the main \textit{Scan} results, with the full \textit{Reason} and \textit{Verify} results provided in Appendix~\ref{app:appendix_b}. Based on these results, we make 5 observations.

\textbf{Current MLLMs remain generally weak.} Although Gemini 3 Pro is among the strongest baselines, it only reaches 60.0 on \textit{Reason}, near the passing threshold, and remains far from reliable on \textit{Scan} (24.3), revealing a persistent limitation of current MLLMs in constructing and reasoning over global evidence views.

\textbf{Evidence removal creates a sharp capability gap.} Even Gemini 3 Pro, the strongest baseline on Reason, only barely achieves a passing threshold on the \textit{Reason} task, while performing considerably worse on \textit{Scan}. This reveals fundamental limitations in current models and training paradigms when it comes to constructing and reasoning over global evidence views. The two instruction-tuned versions of Qwen3-VL perform especially poorly, scoring below 10 and 20 points on \textit{Verify} and \textit{Scan} respectively, highlighting the difficulty of these higher-level tasks.

\textbf{Post-training leads to significant performance improvements.} The SFT- and RL-finetuned variants of Qwen3-VL-8B demonstrate consistent overall improvements across all rewards. A similar pattern of improvement is observed for the Qwen3-VL-235B-A22B series. Our RL-trained model achieves substantial improvements on both \textit{Verify} and \textit{Scan}. It approaches the Qwen3-VL-235B-A22B-Thinking on $R_{\text{completeness}}$ and $R_{\text{alignment}}$, surpasses it in terms of the overall composite score, and further narrows the gap to Gemini 3 Pro. In particular, SFT helps regularize the output through basic format and distributional constraints, leading to a noticeable increase in $R_{\text{precision}}$. This improvement is still evident in the RL phase, where the reward metrics continue to grow, indicating that all 3 metrics play a crucial role in the final performance.

\textbf{VERA-RL shows measurable transfer to ScholScan.} We further evaluate on ScholScan (Table \ref{ScholScan}), and the results mirror those observed above. Notably, although absolute scores remain low in this harder cross-benchmark setting, VERA-RL moves Qwen3-VL-8B from near-zero performance to non-trivial scores and brings it close to Qwen3-VL-235B-A22B-Instruct on several metrics. This suggests that the learned \textit{Reason}--\textit{Verify}--\textit{Scan} capability is not limited to VERA-13K. More details and analysis are provided in Appendix~\ref{app:appendix_c}.

\textbf{\textit{Reason} provides a reference for the internal-knowledge upper bound.} We observe that \textit{Reason} shows relatively limited improvement during post-training, while \textit{Verify} approaches \textit{Reason}.
We consider \textit{Reason} and $R_{\text{completeness}}$ to be more strongly shaped by the model's parametric prior knowledge, making them useful reference points for the upper bound of capability under internal knowledge alone. In contrast, \textit{Verify} and \textit{Scan} demand stronger external evidence grounding and higher-level verifiable reasoning, thus reflecting a shift in the nature of capability from internal recall to evidence-based inference.

\subsection{Additional Analysis}
\begin{figure*}[t]
  \centering
  \begin{minipage}[c]{0.60\linewidth}
    \centering
    \includegraphics[width=\linewidth]{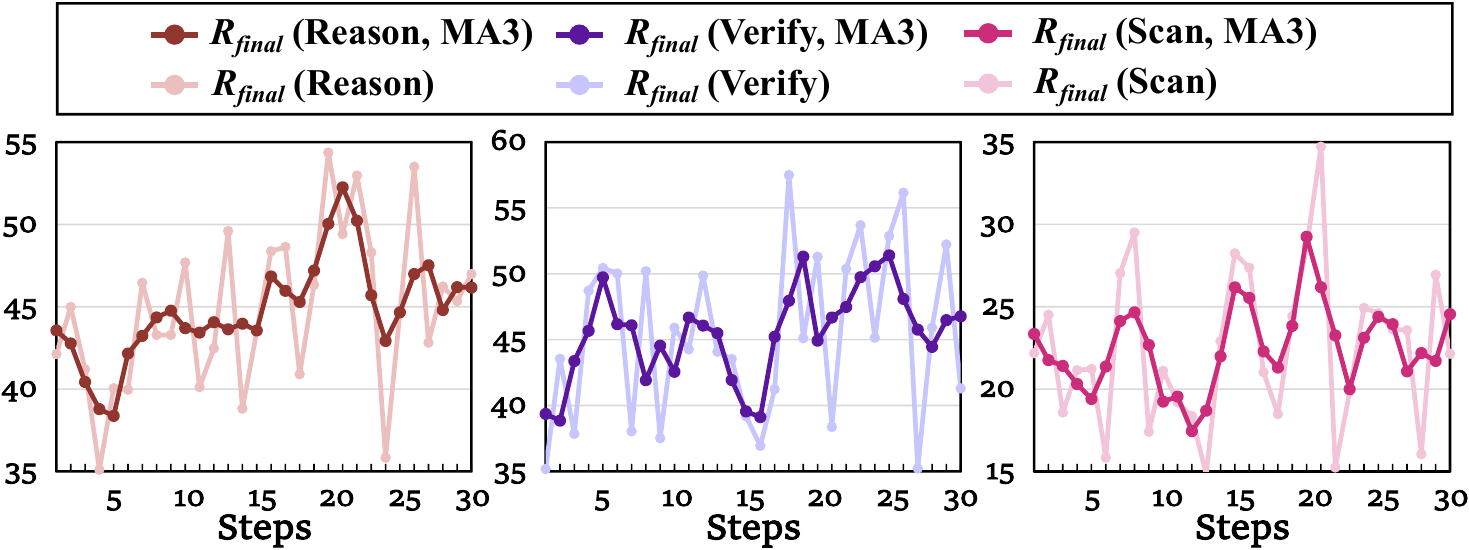}
  \end{minipage}
  \quad
  \begin{minipage}[c]{0.34\linewidth}
    \centering
    \small
    \setlength{\tabcolsep}{4pt}
    \renewcommand{\arraystretch}{1.12}
    \begin{tabular}{@{}ccccc@{}}
      \toprule
      & \textbf{$R_{\text{final}}$} & \textbf{$R_{\text{completeness}}$} & \textbf{$R_{\text{alignment}}$} & \textbf{$R_{\text{precision}}$} \\
      \midrule
      $\uparrow$   & 285 & 40  & 36  & 287 \\
      $\downarrow$ & 6   & 5   & 4   & 3   \\
      $=$          & 9   & 255 & 260 & 10  \\
      Total          & 300 & 300 & 300 & 300  \\
      \bottomrule
    \end{tabular}
  \end{minipage}
  \caption{\textbf{Left}: Training dynamics of main experiments. \textbf{Right}: Metrics change after \textsc{VERA-RL} on \textit{Scan} (\textit{test} set).}
  \label{fig:figure3}
\end{figure*}
\textbf{Training Dynamics.} Figure~\ref{fig:figure3} shows the rollout dynamics of the main experiment. Rewards increase in phases with moderate oscillations, indicating that the staged tasks provide usable optimization signals despite the long and structured outputs.  Additionally, \textit{Reason} and \textit{Verify} exhibit converging patterns in the fine-grained perspective of a single training run. This is consistent with our observation that \textit{Reason} serves as an internal-knowledge reference point.

\textbf{The improvements also vary across error categories.} 
For RQD, PD, and SG, post-training brings larger gains than parameter scaling. These error types often involve domain-specific concepts introduced within the paper, reducing reliance on the model's parametric knowledge and shifting the demand toward evidence-grounded verification. 
DI and IC yield limited improvements, suggesting that errors involving experimental design and final conclusions require stronger holistic understanding and a higher degree of internal scientific knowledge. 
For QI, the model shows broad improvements under both paths, which we attribute to its additional demand on numerical computation and analysis.

\subsection{Ablation Study of Reward Design}
\begin{figure}[t]
  \centering
  \includegraphics[width=\linewidth]{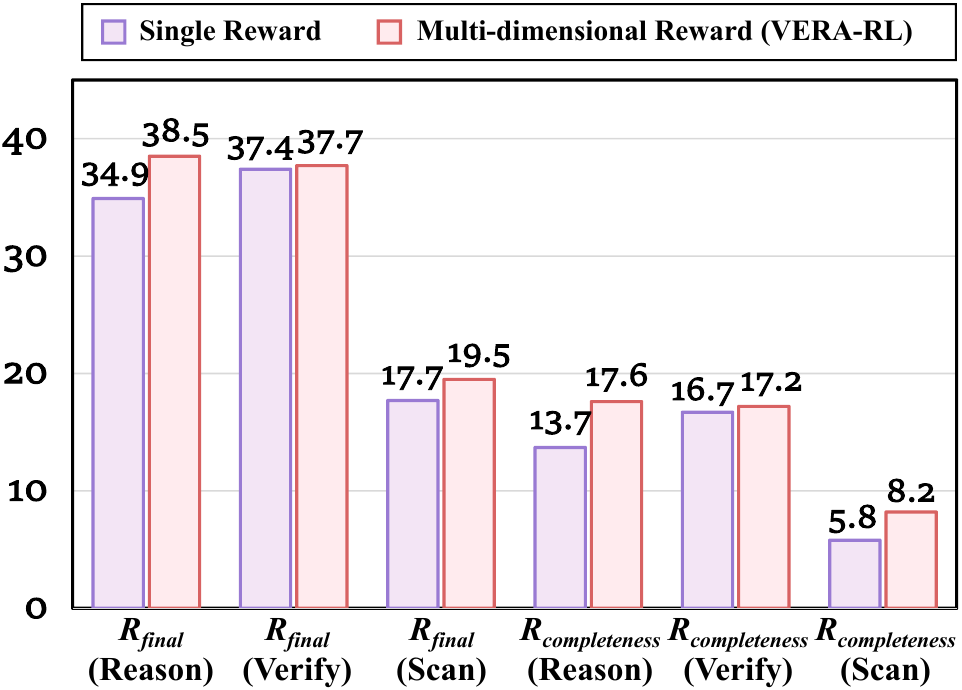}
  \caption{Performance comparison between multi-dimensional and single reward configurations.}
  \label{fig:Figure4}
\end{figure}
\begin{figure}[t]
  \centering
  \includegraphics[width=\linewidth]{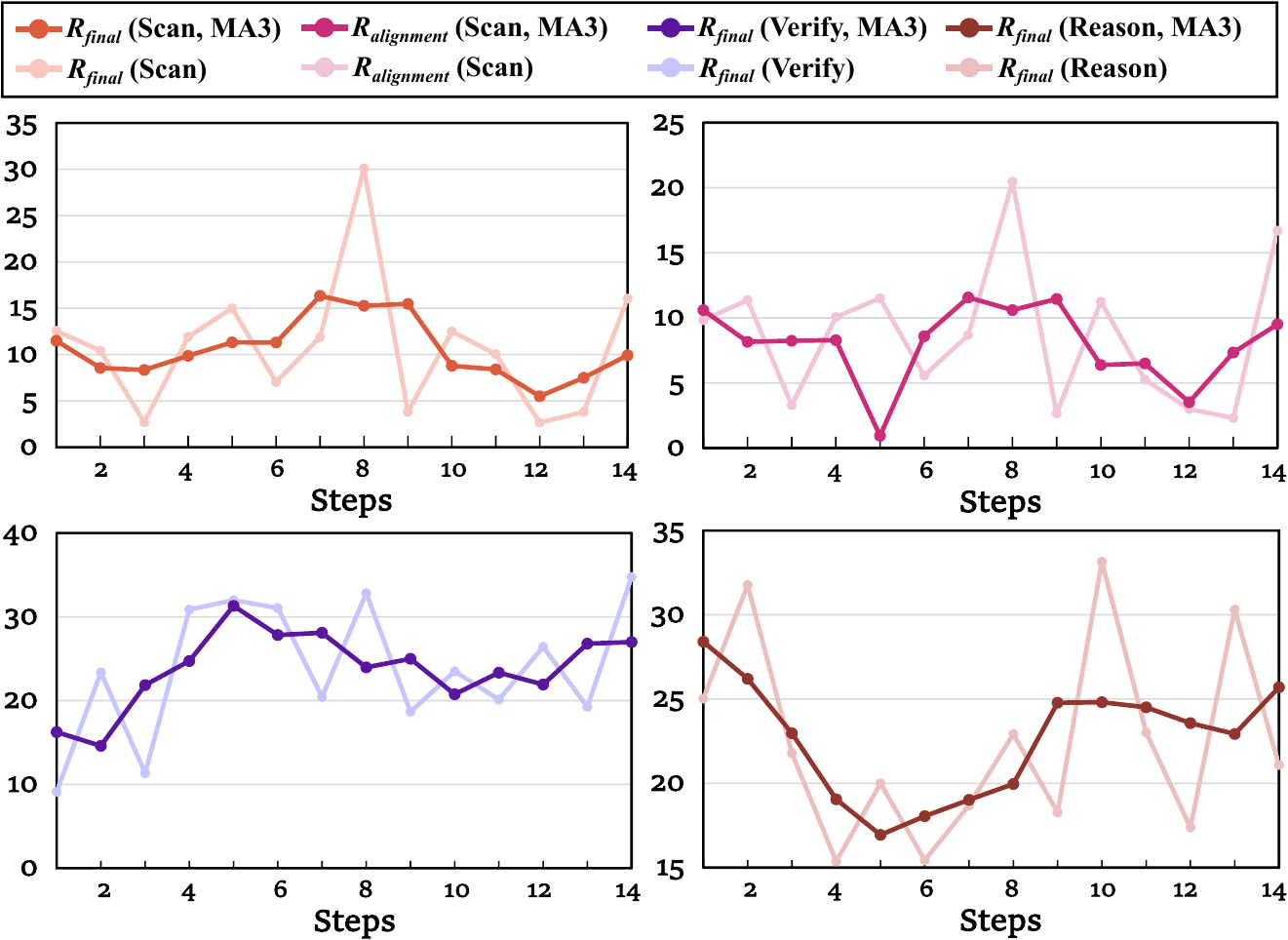}
  \caption{Training dynamics of the ablation study using only \(R_{\mathrm{completeness}}\).}
  \label{fig:Figure5}
\end{figure}

 To explore the necessity of our multi-dimensional reward system, we compare it with a single-reward variant using only $R_{\text{completeness}}$. As shown in Figure~\ref{fig:Figure4}, the multi-dimensional reward consistently outperforms the single-reward variant on both $R_{\text{final}}$ and $R_{\text{completeness}}$. Furthermore, the training dynamics in Figure \ref{fig:Figure5} reveal that reverting to such a sparse reward signal leads to degradation in overall performance. From a task-paradigm perspective, the model shows little meaningful improvement on \textit{Reason} and \textit{Verify}, while its performance on the \textit{Scan} task collapses. More notably, from the perspective of reward optimization itself, the model struggles to stably improve even the explicitly rewarded metric, with training exhibiting clear signs of instability and regression. The reward components are not independent. Instead, they form a tightly coupled positive-feedback structure that collectively and precisely defines the capability dimensions required by the task.

\subsection{Ablation Study of Task Framework}
\begin{table*}[t]
\caption{
Performance comparison across different configurations for \textit{Reason}, \textit{Verify}, and \textit{Scan} tasks. The metric $R_{\text{com}}$ refers to $R_{\text{completeness}}$ and $R_{\text{align}}$ refers to $R_{\text{alignment}}$. The \textit{Pure Scan} configuration isolates the \textit{Scan} task for training.
}
\centering
\setlength{\tabcolsep}{6pt}
\renewcommand{\arraystretch}{1}

\small

\begin{tabular}{cccccccccccc}
\toprule
\textbf{\textit{Config.}} & \textbf{Step} & \multicolumn{3}{c}{\textbf{\textit{Reason}}} & \multicolumn{3}{c}{\textbf{\textit{Verify}}} & \multicolumn{4}{c}{\textbf{\textit{Scan}}} \\
\cmidrule(r){3-5} \cmidrule(r){6-8} \cmidrule(r){9-12}
                  &               & $R_{\text{final}}$ & $R_{\text{com}}$ & $R_{\text{precision}}$ & $R_{\text{final}}$ & $R_{\text{com}}$ & $R_{\text{precision}}$ & $R_{\text{final}}$ & $R_{\text{com}}$ & $R_{\text{align}}$ & $R_{\text{precision}}$ \\
\midrule
Main    & 30 & 38.5 & 17.6 & 69.9 & 37.7 & 17.2 & 68.5 & 19.5 & 8.2 & 6.2 & 68.6 \\
Main    & 15 & 36.1 & 14.6 & 68.4 & 35.0 & 14.9 & 65.2 & 15.0 & 2.8 & 2.3 & 65.0 \\
Pure Scan & 10 & 37.0 & 16.7 & 67.4 & 34.0 & 12.8 & 65.8 & 16.8 & 5.3 & 3.7 & 66.0 \\
\rowcolor{gray!10} 
\textit{$\Delta$ to Main}(30) & - & \textcolor{blue}{-1.5} & \textcolor{blue}{-0.9} & \textcolor{blue}{-2.5} & \textcolor{blue}{-3.7} & \textcolor{blue}{-4.4} & \textcolor{blue}{-2.7} & \textcolor{blue}{-2.7} & \textcolor{blue}{-2.9} & \textcolor{blue}{-2.5} & \textcolor{blue}{-2.6} \\

Pure Scan & 20 & 37.2 & 15.9 & 69.1 & 34.1 & 12.6 & 66.3 & 16.9 & 5.3 & 4.2 & 65.4\\
\rowcolor{gray!10} 
\textit{$\Delta$ to Main}(30) & - & \textcolor{blue}{-1.3} & \textcolor{blue}{-1.7} & \textcolor{blue}{-0.8} & \textcolor{blue}{-3.6} & \textcolor{blue}{-4.6} & \textcolor{blue}{-2.2} & \textcolor{blue}{-2.6} & \textcolor{blue}{-2.9} & \textcolor{blue}{-2.0} & \textcolor{blue}{-3.2} \\
Curricular & 20 & 36.3 & 14.3 & 69.2 & 35.8 & 15.1 & 66.8 & 18.2 & 6.8 & 5.2 & 67.3 \\

\rowcolor{gray!10} 
\textit{$\Delta$ to Main}(30) & - & \textcolor{blue}{-2.2} & \textcolor{blue}{-3.3} & \textcolor{blue}{-0.7} & \textcolor{blue}{-1.9} & \textcolor{blue}{-2.1} & \textcolor{blue}{-1.7} & \textcolor{blue}{-1.3} & \textcolor{blue}{-1.4} & \textcolor{blue}{-1.0} & \textcolor{blue}{-1.3} \\
Curricular & 30 & 36.5 & 15.5 & 67.9 & 36.1 & 15.1 & 67.6 & 16.5 & 4.5 & 4.2 & 64.8\\

\rowcolor{gray!10} 
\textit{$\Delta$ to Main}(30) & - & \textcolor{blue}{-2.0} & \textcolor{blue}{-2.1} & \textcolor{blue}{-2.0} & \textcolor{blue}{-1.6} & \textcolor{blue}{-2.1} & \textcolor{blue}{-0.9} & \textcolor{blue}{-3.0} & \textcolor{blue}{-3.7} & \textcolor{blue}{-2.0} & \textcolor{blue}{-3.8} \\
\bottomrule
\end{tabular}
\label{scan_only}
\end{table*}
\begin{figure}[t]
  \centering
  \includegraphics[width=\linewidth]{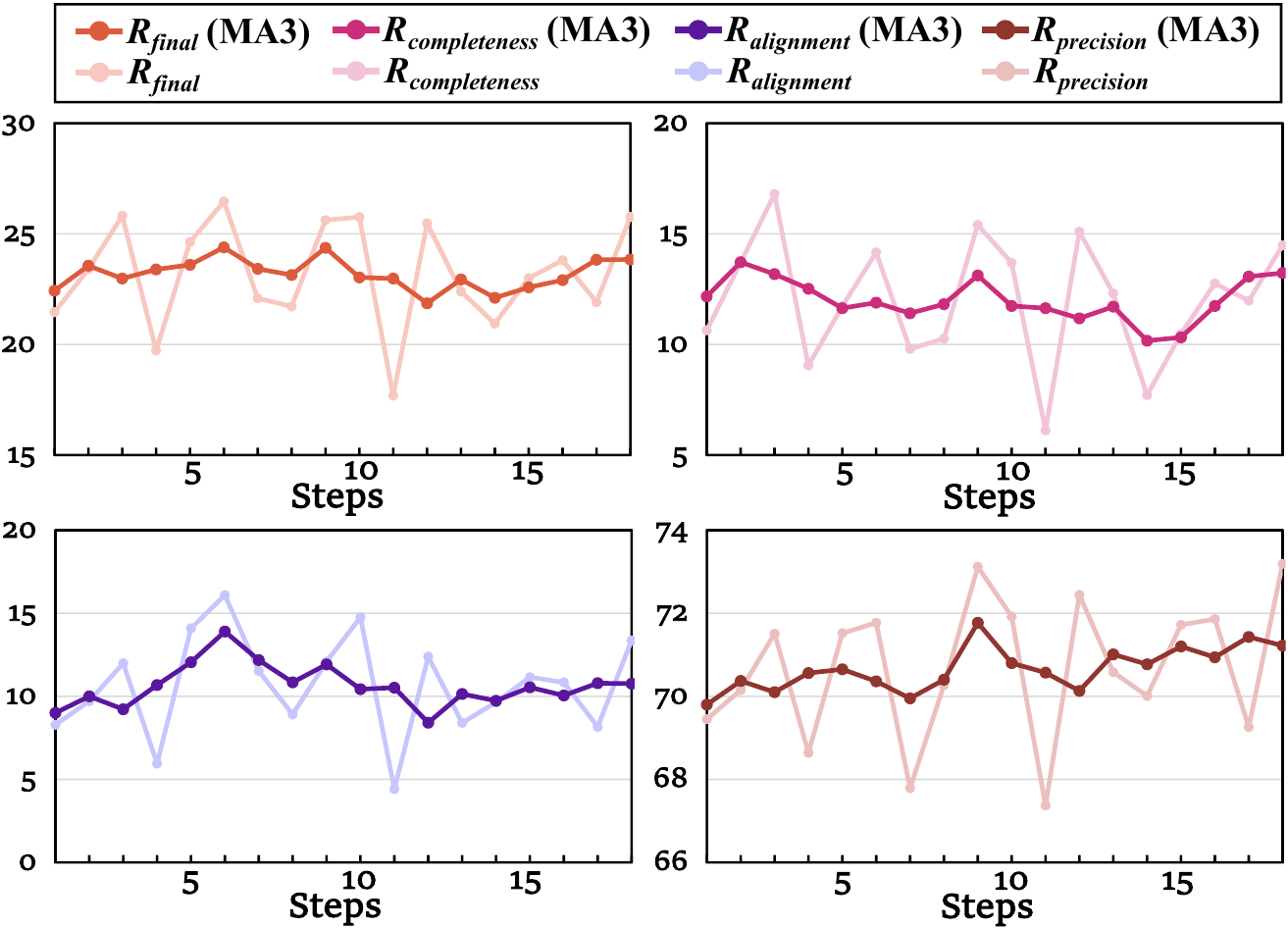}
  \caption{Training dynamics of the ablation experiment using only \textit{Scan} tasks.}
  \label{fig:Figure6}
\end{figure}

To evaluate the effectiveness of the three-stage task paradigm, we compare the main setting with two alternatives in Table~\ref{scan_only}. The first setting, \textit{Pure Scan}, trains the model only on \textit{Scan} samples. When controlling for the amount of \textit{Scan} exposure at step 10, \textit{Pure Scan} performs worse than the main setting not only on \textit{Reason} and \textit{Verify}, but also on \textit{Scan} itself. Extending \textit{Pure Scan} training to step 20 still fails to close the gap. Just as the reward components are complementary, the task stages are also coupled in training. \textit{Reason} and \textit{Verify} remain non-trivial, and removing their supervision weakens the evidence-conditioned reasoning and judgment needed for \textit{Scan}. Neglecting the training of these tasks may hinder the development of \textit{Scan} performance.

Further training does not yield stable improvement on \textit{Reason}, and \textit{Scan} also improves slowly. Figure \ref{fig:Figure6} illustrates the reward dynamics when training solely on the \textit{Scan} task, which reveal the struggle across nearly all metrics from a fine-grained perspective.

Furthermore, we introduced an attempt with curricular learning~\cite{yuan2025vlcogitoprogressivecurriculumreinforcement}. Specifically, we used the main experiment setup for the first 15 steps, then switched to training using only \textit{Scan} samples, and evaluated models at total steps 20 and 30. The results in Table \ref{scan_only} indicate that this setup faces similar challenges as the setup using only \textit{Scan} task samples. This further underscores the tight coupling of our task paradigm framework. From a more practical perspective, it enables the expansion of sample size by breaking down capability dimensions, efficiently utilizing valuable high-quality papers, and reserving greater potential for scaling up.

\subsection{Case Study}
We further examine \textit{Scan} samples improved by \textsc{VERA-RL}. Statistics in Figure~\ref{fig:figure3} show broad gains across many instances, and the cases are provided in Appendix~\ref{app:appendix_f}. 
Qualitatively, the base Qwen3-VL-8B-Instruct often stops at surface-level consistency checks and gives broad no-error judgments. 
After \textsc{VERA-RL}, the model searches more actively across the paper, identifies localized evidence from dispersed content, and connects it into traceable error reasoning. 
These cases suggest that the gains are not merely from catering to evaluator heuristics, but reflect stronger full-paper exploration and evidence-grounded verification required by \textit{Scan}.

\section{Conclusion}
We presented \textsc{VERA-RL} for training verifiable scientific error detection over academic papers. By decomposing \textit{Scan} into \textit{Reason}, \textit{Verify}, and \textit{Scan}, and by using rewards for reasoning completeness, evidence alignment, and error precision, \textsc{VERA-RL} provides a principled training path from evidence-specified reasoning to issue- and evidence-absent verification. We also constructed \textsc{VERA-13K}, a 12,900-sample dataset of matched three-stage chains across 6 scientific-error categories. Experiments on \textsc{VERA-13K} and ScholScan show consistent improvements, and ablations confirm that both task staging and reward design are necessary for stable gains. These results demonstrate that \textit{Scan}-style verification is a trainable capability that can be systematically improved through aligned task design, data construction, and reward modeling, offering a concrete step from scientific assistance toward more autonomous scientific research.

\section*{Limitations}
This work focuses on scientific error detection as a concrete setting for Scan-style verification over academic papers. It does not cover the full range of peer-review judgments, such as novelty, significance, writing quality, or broader research impact, which often depend on community context and subjective assessment.
\textsc{VERA-13K} emphasizes errors that can be verified from the paper itself, making supervision and reward computation more structured. As a result, more implicit weaknesses or errors requiring extensive external domain knowledge may be underrepresented.
Our training experiments are mainly conducted on Qwen3-VL-8B. Although we compare with multiple strong MLLMs and external benchmarks, larger-scale RL training, broader model families, and alternative paper representations remain for future study.

\section*{Acknowledgments}
This work is supported by WeChat AI, Tencent Inc., China and ``The Fundamental Research Funds for the Central Universities, Peking University''. We would also like to thank the anonymous reviewers and area chairs for constructive discussions and feedback.

\bibliography{custom}

\appendix
\clearpage
\section{Details of \textsc{VERA-13K}}
\label{app:appendix_a}

\paragraph{Comparison with other benchmarks} Table~\ref{tab:benchmark_comparison} summarizes the differences in modality, task paradigm, etc. with existing academic document understanding benchmarks.

\begin{table*}[t]
\centering
\caption{Comparison with representative academic document understanding benchmarks. \textbf{T}: text; \textbf{I}: image; \textbf{MD}: multimodal document; \textbf{CS}: Computer Science.}
%\small
\label{tab:benchmark_comparison}
%\resizebox{\columnwidth}{!}
{
\begin{tabular}{lccccc}
\toprule
\textbf{Benchmark} & \textbf{Modality} & \textbf{Paradigm} & \textbf{Eval.} & \textbf{Domains} & \textbf{Count (K)} \\
\midrule
CharXiv       & I      & Reason             & Close & 8  & 11.6 \\
ArXivQA       & I      & Reason             & Close & 10 & 100  \\
MMCR          & T+MD   & Reason             & Close & CS & 310  \\
AAAR-1.0      & T+MD   & Reason             & Mixed & CS & 13.5 \\
PRISMM-Bench  & I      & Reason             & Close & CS & 0.4  \\
ScholScan     & T+MD   & Scan               & Open  & 13 & 1.8  \\
\textbf{\textsc{VERA-13K} (ours)} & \textbf{T+MD} & \textbf{Reason+Verify+Scan} & \textbf{Open}  & \textbf{9}  & \textbf{12.9} \\
\bottomrule
\end{tabular}
}
\end{table*}

\begin{figure}[t]
  \centering
  \includegraphics[width=0.8\linewidth]{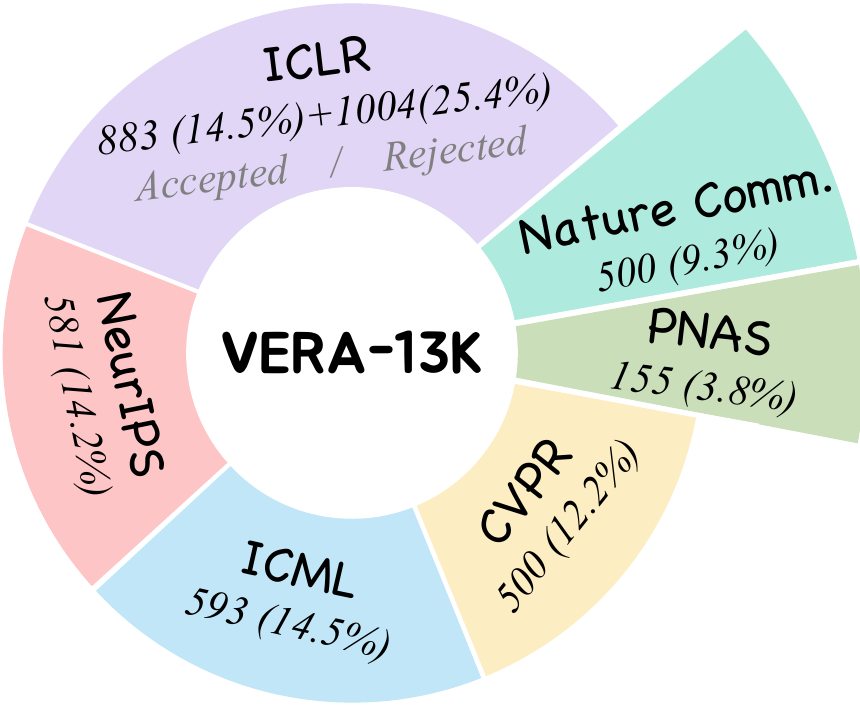}
  \caption{Distribution of paper sources.}
  \label{fig:distribution}
\end{figure}

\paragraph{Distribution} Figure~\ref{fig:distribution} presents the source distribution of papers in \textsc{VERA-13K}. Table~\ref{tab:error_type_distribution} reports the distribution of the 6 categories in the \textit{train} subset. The categories cover typical risk points across the research process, from problem formulation and experimental design to quantitative analysis and final inference.

AI/ML papers are used as a major source because they represent a rapidly growing and highly active research area with abundant public materials and reviews. Also, they contain diverse subfields. We further categorize them according to the ICML call-for-papers taxonomy and report the major represented subfields in Table~\ref{tab:ai_subdomain_distribution}.

Beyond computer science, \textsc{VERA-13K} includes 655 papers and 1,605 samples from broader scientific domains. We group these papers into high-level scientific domains. As shown in Table~\ref{tab:broad_domain_distribution}, \textsc{VERA-13K} covers diverse areas beyond computer science.

\textbf{Source Composition.}
VERA-13K combines two complementary data sources: controlled-edit samples constructed from accepted papers and review-derived samples based on objective errors extracted from peer reviews. Table~\ref{tab:source_composition} reports their proportions in each split. Review-derived samples account for 27.2\%, 32.8\%, and 30.7\% of the SFT, RL, and test sets, respectively, yielding a similar source composition across training and evaluation splits.

\begin{table}[t]
\centering
%\small
\caption{Broad-domain distribution beyond Computer Science.}
\label{tab:broad_domain_distribution}
\begin{tabular}{lr}
\toprule
\textbf{Domain} & \textbf{Count} \\
\midrule
Life Sciences      & 227 \\
Medicine           & 107 \\
Materials          & 67  \\
Ecology            & 34  \\
Chemistry          & 32  \\
Environment        & 30  \\
Physics            & 23  \\
Interdisciplinary  & 135 \\
\midrule
Total              & 655 \\
\bottomrule
\end{tabular}
\end{table}

\begin{table}[t]
\centering
%\small
\caption{Major AI/ML subdomain distribution. Counts are paper-level statistics, and only the most represented subfields are listed.}
\label{tab:ai_subdomain_distribution}
\begin{tabular}{lr}
\toprule
\textbf{Subdomain} & \textbf{Count} \\
\midrule
Foundation Models \& LLM & 424 \\
Computer Vision          & 412 \\
Multimodal Learning      & 287 \\
Generative Modeling      & 243 \\
Reinforcement Learning   & 194 \\
Evaluation               & 190 \\
Graph ML                 & 162 \\
Safety                   & 156 \\
Learning Theory          & 119 \\
AI Agents                & 115 \\
Efficiency               & 94  \\
\bottomrule
\end{tabular}
\end{table}

\begin{table}[t]
\centering
%\small
\caption{Train-subset distribution across 6 scientific-error categories.}
\label{tab:error_type_distribution}
\begin{tabular}{lr}
\toprule
\textbf{Category} & \textbf{Count} \\
\midrule
QI  & 3450 \\
DI  & 540  \\
IC  & 2190 \\
PD  & 1470 \\
RQD & 3450 \\
SG  & 900  \\
\bottomrule
\end{tabular}
% TODO: Verify these counts before final submission. The current rebuttal numbers sum to 12,060, while the main training split is reported as 12,000 samples.
\end{table}

\begin{table*}[t]
\centering
\caption{Source composition of VERA-13K. Review-derived samples are constructed from objective errors extracted from peer reviews, while controlled-edit samples are generated by injecting errors into accepted papers.}
\begin{tabular}{lrrrr}
\toprule
Split & Total & Review-derived & Controlled-edit & Review (\%) \\
\midrule
SFT Train & 10,500 & 2,856 & 7,644 & 27.2 \\
RL Train  & 1,500  & 492   & 1,008 & 32.8 \\
Test      & 900    & 276   & 624   & 30.7 \\
\midrule
Total     & 12,900 & 3,624 & 9,276 & 28.1 \\
\bottomrule
\end{tabular}
\label{tab:source_composition}
\end{table*}

\section{Supplementary Materials of Experiments}
\label{app:appendix_b}
\newcommand{\SuppScanMetricBlock}[1]{%
  \rowcolor{gray!10}%
  \multicolumn{8}{c}{\textbf{Scan}\quad $#1$}\\[-0.05ex]
}
\subsection{Supplementary Results}
Table~\ref{tab:scan_supp_metrics} provides supplementary \textit{Scan} sub-metrics for additional baseline models. Tables~\ref{tab:reason_detailed_metrics} and~\ref{tab:verify_detailed_metrics} report the full \textit{Reason} and \textit{Verify} results across error categories. For \textit{Reason} and \textit{Verify}, we report $R_{\mathrm{completeness}}$, $R_{\mathrm{precision}}$, and $R_{\mathrm{final}}$, since evidence alignment is disabled in these evidence-specified settings. All scores are scaled by 100.

\begin{table*}[t]
\caption{Supplementary sub-metric scores on the \textit{Scan} task (scaled by 100).}
\label{tab:scan_supp_metrics}
\centering
{%
\small
\setlength{\tabcolsep}{5pt}
\setlength{\extrarowheight}{0.25ex}
\renewcommand{\arraystretch}{1.08}
\begin{tabular}{>{\raggedright\arraybackslash}p{4.6cm} *{7}{>{\centering\arraybackslash}p{1cm}}}
\toprule
\textbf{Models} & \textbf{Avg.} & \textbf{QI} & \textbf{DI} & \textbf{IC} & \textbf{PD} & \textbf{RQD} & \textbf{SG} \\
\midrule

\SuppScanMetricBlock{R_{\mathrm{completeness}}}
GPT-5.4 & 39.5 & 54.3 & 42.6 & 33.0 & 36.8 & 37.7 & 32.9 \\
Qwen3-VL-Plus & 20.6 & 34.2 & 7.5 & 28.5 & 16.0 & 19.2 & 18.3 \\
Qwen3-VL-32B & 12.7 & 9.0 & 6.7 & 19.5 & 10.0 & 13.8 & 17.3 \\
Seed-1.6-Thinking & 13.4 & 19.1 & 11.2 & 15.7 & 8.8 & 14.0 & 11.5 \\

\SuppScanMetricBlock{R_{\mathrm{alignment}}}
GPT-5.4 & 28.5 & 60.0 & 17.0 & 19.5 & 37.5 & 23.1 & 16.7 \\
Qwen3-VL-Plus & 15.3 & 26.2 & 6.1 & 21.4 & 11.8 & 12.7 & 13.9 \\
Qwen3-VL-32B & 10.1 & 12.1 & 5.0 & 16.6 & 7.3 & 9.1 & 10.3 \\
Seed-1.6-Thinking & 11.1 & 20.1 & 8.1 & 14.7 & 6.6 & 9.2 & 8.2 \\

\SuppScanMetricBlock{R_{\mathrm{precision}}}
GPT-5.4 & 13.7 & 20.6 & 11.5 & 9.1 & 10.4 & 9.7 & 19.9 \\
Qwen3-VL-Plus & 59.2 & 57.9 & 61.4 & 50.1 & 54.8 & 58.9 & 71.9 \\
Qwen3-VL-32B & 60.2 & 63.1 & 56.2 & 58.5 & 67.7 & 58.1 & 57.7 \\
Seed-1.6-Thinking & 56.7 & 51.4 & 47.0 & 77.0 & 49.6 & 66.8 & 48.6 \\

\bottomrule
\end{tabular}
}
\end{table*}

\newcommand{\RVMetricBlock}[1]{%
  \rowcolor{gray!10}%
  \multicolumn{8}{c}{\textbf{#1}}\\[-0.05ex]
}

\begin{table*}[t]
\caption{Sub-metric scores on the \textit{Reason} task (scaled by 100).}
\label{tab:reason_detailed_metrics}
\centering
{%
\small
\setlength{\tabcolsep}{5pt}
\setlength{\extrarowheight}{0.25ex}
\renewcommand{\arraystretch}{1.08}
\begin{tabular}{>{\raggedright\arraybackslash}p{5cm} *{7}{>{\centering\arraybackslash}p{1cm}}}
\toprule
\textbf{Models} & \textbf{Avg.} & \textbf{QI} & \textbf{DI} & \textbf{IC} & \textbf{PD} & \textbf{RQD} & \textbf{SG} \\
\midrule

\RVMetricBlock{\textit{Reason}\quad $R_{\mathrm{completeness}}$}
GPT-5.4 & 71.1 & 82.1 & 62.8 & 58.4 & 78.6 & 72.0 & 71.9 \\
Gemini 3 Pro & 59.2 & 64.7 & 55.3 & 63.0 & 57.0 & 65.2 & 49.9 \\
Seed-1.6-Thinking & 52.4 & 58.4 & 49.8 & 52.0 & 58.4 & 59.6 & 36.3 \\
Qwen3-VL-Plus & 60.7 & 68.8 & 52.3 & 68.0 & 60.8 & 64.8 & 49.6 \\
Qwen3-VL-235B-A22B (Thinking) & 54.7 & 67.1 & 42.7 & 64.0 & 58.1 & 55.2 & 40.9 \\
Qwen3-VL-235B-A22B (Instruct) & 46.8 & 52.4 & 36.1 & 52.4 & 49.5 & 49.9 & 40.4 \\
Qwen3-VL-32B & 52.2 & 63.7 & 45.4 & 46.2 & 61.6 & 53.5 & 42.8 \\
Qwen3-VL-8B (Instruct) & 23.3 & 25.3 & 17.3 & 27.2 & 26.6 & 27.3 & 15.9 \\
Qwen3-VL-8B (SFT) & 13.9 & 16.1 & 13.6 & 4.7 & 20.6 & 19.5 & 8.9 \\
Qwen3-VL-8B (SFT+RL) & 17.6 & 12.7 & 9.8 & 17.3 & 24.2 & 24.0 & 17.6 \\

\RVMetricBlock{\textit{Reason}\quad $R_{\mathrm{precision}}$}
GPT-5.4 & 34.4 & 40.8 & 28.3 & 26.0 & 44.7 & 43.9 & 24.0 \\
Gemini 3 Pro & 61.2 & 63.1 & 57.6 & 58.0 & 65.0 & 67.1 & 56.6 \\
Seed-1.6-Thinking & 49.8 & 55.6 & 44.3 & 48.2 & 50.7 & 56.5 & 43.7 \\
Qwen3-VL-Plus & 53.3 & 58.1 & 50.8 & 50.9 & 62.8 & 53.0 & 44.1 \\
Qwen3-VL-235B-A22B (Thinking) & 47.4 & 58.2 & 39.7 & 41.4 & 53.5 & 51.5 & 39.9 \\
Qwen3-VL-235B-A22B (Instruct) & 44.5 & 55.2 & 42.2 & 39.1 & 49.6 & 44.7 & 36.3 \\
Qwen3-VL-32B & 44.8 & 54.3 & 40.2 & 38.2 & 49.4 & 47.3 & 39.4 \\
Qwen3-VL-8B (Instruct) & 38.8 & 40.9 & 38.2 & 42.1 & 46.8 & 38.1 & 26.8 \\
Qwen3-VL-8B (SFT) & 54.9 & 53.8 & 53.6 & 51.6 & 58.4 & 58.6 & 53.3 \\
Qwen3-VL-8B (SFT+RL) & 69.9 & 68.1 & 66.3 & 71.3 & 71.7 & 71.9 & 69.9 \\

\RVMetricBlock{\textit{Reason}\quad $R_{\mathrm{final}}$}
GPT-5.4 & 56.4 & 65.6 & 49.0 & 45.4 & 65.0 & 60.7 & 52.7 \\
Gemini 3 Pro & 60.0 & 64.0 & 56.2 & 61.0 & 60.2 & 66.0 & 52.6 \\
Seed-1.6-Thinking & 51.4 & 57.3 & 47.6 & 50.5 & 55.3 & 58.4 & 39.3 \\
Qwen3-VL-Plus & 57.7 & 64.5 & 51.7 & 61.2 & 61.6 & 60.1 & 47.4 \\
Qwen3-VL-235B-A22B (Thinking) & 51.8 & 63.5 & 41.5 & 55.0 & 56.3 & 53.7 & 40.5 \\
Qwen3-VL-235B-A22B (Instruct) & 45.9 & 53.5 & 38.5 & 47.1 & 49.5 & 47.8 & 38.8 \\
Qwen3-VL-32B & 49.2 & 59.9 & 43.3 & 43.0 & 56.7 & 51.0 & 41.5 \\
Qwen3-VL-8B (Instruct) & 29.5 & 31.5 & 25.7 & 33.2 & 34.7 & 31.6 & 20.3 \\
Qwen3-VL-8B (SFT) & 30.3 & 31.2 & 29.6 & 23.5 & 35.7 & 35.1 & 26.6 \\
Qwen3-VL-8B (SFT+RL) & 38.5 & 34.8 & 32.4 & 38.9 & 43.2 & 43.2 & 38.5 \\

\bottomrule
\end{tabular}
}
\end{table*}

\begin{table*}[t]
\caption{Sub-metric scores on the \textit{Verify} task (scaled by 100).}
\label{tab:verify_detailed_metrics}
\centering
{%
\small
\setlength{\tabcolsep}{5pt}
\setlength{\extrarowheight}{0.25ex}
\renewcommand{\arraystretch}{1.08}
\begin{tabular}{>{\raggedright\arraybackslash}p{5cm} *{7}{>{\centering\arraybackslash}p{1cm}}}
\toprule
\textbf{Models} & \textbf{Avg.} & \textbf{QI} & \textbf{DI} & \textbf{IC} & \textbf{PD} & \textbf{RQD} & \textbf{SG} \\
\midrule

\RVMetricBlock{\textit{Verify}\quad $R_{\mathrm{completeness}}$}
GPT-5.4 & 62.3 & 74.7 & 51.2 & 67.6 & 65.5 & 71.0 & 49.7 \\
Gemini 3 Pro & 52.3 & 68.8 & 40.5 & 46.3 & 62.4 & 55.0 & 40.5 \\
Seed-1.6-Thinking & 32.8 & 37.3 & 22.0 & 27.2 & 39.9 & 42.2 & 28.4 \\
Qwen3-VL-Plus & 46.4 & 64.2 & 36.7 & 43.0 & 52.1 & 46.9 & 35.7 \\
Qwen3-VL-235B-A22B (Thinking) & 41.0 & 57.6 & 15.8 & 42.0 & 50.3 & 47.5 & 32.5 \\
Qwen3-VL-235B-A22B (Instruct) & 13.0 & 14.2 & 9.0 & 16.0 & 16.2 & 13.0 & 9.3 \\
Qwen3-VL-32B & 40.7 & 53.6 & 29.5 & 49.4 & 41.0 & 42.2 & 28.7 \\
Qwen3-VL-8B (Instruct) & 3.9 & 4.3 & 5.5 & 5.0 & 3.3 & 3.7 & 1.5 \\
Qwen3-VL-8B (SFT) & 12.6 & 11.8 & 5.5 & 15.5 & 9.8 & 19.7 & 13.1 \\
Qwen3-VL-8B (SFT+RL) & 17.2 & 14.9 & 9.2 & 20.3 & 22.9 & 18.3 & 17.7 \\

\RVMetricBlock{\textit{Verify}\quad $R_{\mathrm{precision}}$}
GPT-5.4 & 46.1 & 60.4 & 36.2 & 46.9 & 51.8 & 45.4 & 39.2 \\
Gemini 3 Pro & 57.8 & 64.2 & 46.1 & 54.7 & 69.2 & 57.8 & 55.0 \\
Seed-1.6-Thinking & 79.4 & 77.5 & 82.4 & 78.1 & 79.9 & 80.4 & 77.9 \\
Qwen3-VL-Plus & 70.4 & 70.6 & 65.7 & 73.6 & 74.1 & 73.0 & 65.2 \\
Qwen3-VL-235B-A22B (Thinking) & 55.5 & 70.2 & 33.2 & 49.9 & 62.3 & 65.3 & 52.2 \\
Qwen3-VL-235B-A22B (Instruct) & 16.1 & 19.9 & 7.6 & 21.1 & 22.2 & 13.6 & 12.2 \\
Qwen3-VL-32B & 66.8 & 69.1 & 61.3 & 65.6 & 68.0 & 72.2 & 64.7 \\
Qwen3-VL-8B (Instruct) & 8.5 & 14.4 & 9.1 & 10.9 & 7.0 & 4.2 & 5.3 \\
Qwen3-VL-8B (SFT) & 52.8 & 50.2 & 49.0 & 55.3 & 54.5 & 57.1 & 50.4 \\
Qwen3-VL-8B (SFT+RL) & 68.5 & 67.5 & 66.9 & 70.7 & 72.4 & 67.0 & 66.6 \\

\RVMetricBlock{\textit{Verify}\quad $R_{\mathrm{final}}$}
GPT-5.4 & 55.8 & 69.0 & 45.2 & 59.3 & 60.0 & 60.7 & 45.5 \\
Gemini 3 Pro & 54.5 & 67.0 & 42.7 & 50.0 & 65.1 & 56.1 & 46.3 \\
Seed-1.6-Thinking & 51.4 & 53.4 & 46.2 & 47.5 & 55.9 & 57.5 & 48.2 \\
Qwen3-VL-Plus & 56.0 & 66.8 & 48.3 & 55.2 & 60.9 & 57.4 & 47.5 \\
Qwen3-VL-235B-A22B (Thinking) & 46.8 & 62.7 & 22.8 & 45.2 & 55.1 & 54.5 & 40.3 \\
Qwen3-VL-235B-A22B (Instruct) & 14.2 & 16.5 & 8.4 & 18.0 & 18.6 & 13.2 & 10.5 \\
Qwen3-VL-32B & 51.2 & 59.8 & 42.2 & 55.9 & 51.8 & 54.2 & 43.1 \\
Qwen3-VL-8B (Instruct) & 5.7 & 8.4 & 6.9 & 7.4 & 4.8 & 3.9 & 3.0 \\
Qwen3-VL-8B (SFT) & 28.6 & 27.2 & 22.9 & 31.4 & 27.7 & 34.7 & 28.0 \\
Qwen3-VL-8B (SFT+RL) & 37.7 & 35.9 & 32.3 & 40.5 & 42.7 & 37.8 & 37.3 \\

\bottomrule
\end{tabular}
}
\end{table*}

Beyond ScholScan, we further evaluate the model on three related external benchmarks covering long-document understanding and scientific reasoning. As shown in Table~\ref{tab:external_eval}, VERA-RL maintains or modestly improves performance across all three benchmarks, suggesting that its gains do not come at the cost of related general capabilities.

We further remove test samples whose source papers overlap with the training set and reevaluate the models. As shown in Table~\ref{tab:paper_disjoint}, the results remain nearly unchanged, and the gains from SFT and RL persist under the paper-disjoint setting.

\begin{table}[t]
\centering
\caption{Additional external evaluation results.}
\begin{tabular}{lcc}
\toprule
Benchmark & Instruct & SFT+RL \\
\midrule
MMLongBench-Doc & 22.9 & 23.3 \\
MMMU            & 44.3 & 46.1 \\
PRISMM-Bench    & 51.6 & 52.0 \\
\bottomrule
\end{tabular}
\label{tab:external_eval}
\end{table}

\begin{table}[t]
\centering
\caption{Performance on the full and paper-disjoint test sets. All scores are $R_{\mathrm{final}}$ and scaled by 100.}
\begin{tabular}{llrrr}
\toprule
Model & Split & \textit{Reason} & \textit{Verify} & \textit{Scan} \\
\midrule
\multirow{2}{*}{Instruct}
& Full     & 29.5 & 5.7 & 2.0 \\
& Disjoint & 30.2 & 5.2 & 1.6 \\
\midrule
\multirow{2}{*}{SFT}
& Full     & 30.3 & 28.6 & 14.4 \\
& Disjoint & 31.3 & 28.6 & 14.3 \\
\midrule
\multirow{2}{*}{SFT+RL}
& Full     & 38.5 & 37.7 & 19.5 \\
& Disjoint & 39.3 & 39.1 & 19.3 \\
\bottomrule
\end{tabular}
\label{tab:paper_disjoint}
\end{table}

\subsection{Training Configurations}
\paragraph{SFT Training}
We perform SFT training on Qwen3-VL-8B-Instruct using a standard teacher-forcing objective. Training is conducted for 1 epoch with a global batch size of 8 and a micro-batch size of 1 per GPU. Optimization is performed using the AdamW optimizer with a learning rate of 5$\times10^{-6}$, no weight decay, and a cosine learning rate schedule. A warm-up ratio of 0.03 is applied, and gradient norms are clipped at 1.0. All experiments use a fixed random seed of 42.

\paragraph{RL Training}
We fine-tune the SFT-initialized model using the DAPO setup. The base model is Qwen3-VL-8B initialized from the SFT checkpoint. We sample 8 responses per prompt. Policy optimization is performed with a global training batch size of 32. We use asymmetric clipping with clip ratios $(\epsilon_{\text{low}}, \epsilon_{\text{high}}) = (0.1, 0.5)$ and adopt a token-level loss aggregation strategy (\texttt{token-mean}) to mitigate length bias. The learning rate is set to 2$\times10^{-6}$, with no weight decay and a cosine scheduler with a warm-up ratio of 0.03. Training is run for 30 optimization steps with a fixed random seed of 42. We enable gradient checkpointing, activation offloading, and FSDP with parameter and optimizer offloading.

\subsection{Training Dynamics}

We further evaluate the step-200 SFT checkpoint to test whether the early plateau in training loss is sufficient. As shown in Table~\ref{tab:sft_step200}, step 200 already improves over the Instruct model, but the 1-epoch checkpoint achieves stronger \textit{Scan} completeness and a higher overall \textit{Scan} score, so we use the latter to initialize RL training.

\begin{table}[t]
\caption{Performance of the intermediate SFT checkpoint at step 200, compared with the original base model and the 1-epoch SFT model. Scores are scaled by 100.}
\label{tab:sft_step200}
\centering
\small
\setlength{\tabcolsep}{5pt}
\renewcommand{\arraystretch}{1.05}
\begin{tabular}{lccc}
\toprule
\textbf{Metric} & \textbf{0 step} & \textbf{200 steps} & \textbf{1 epoch} \\
\midrule
Scan $R_{\mathrm{completeness}}$ & 1.5 & 1.6 & 5.3 \\
Scan $R_{\mathrm{alignment}}$    & 1.0 & 2.4 & 2.5 \\
Scan $R_{\mathrm{precision}}$    & 5.0 & 61.4 & 56.3 \\
Scan $R_{\mathrm{final}}$        & 2.0 & 13.8 & 14.4 \\
Verify $R_{\mathrm{final}}$      & 5.7 & 29.2 & 28.6 \\
Reason $R_{\mathrm{final}}$      & 29.5 & 31.1 & 30.3 \\
\bottomrule
\end{tabular}
\end{table}
\section{Metrics Definition in ScholScan}
\label{app:appendix_c}
Given a model answer $a$, ScholScan first parses it into a structured tuple:
\begin{equation*}
\Psi(a) =
\left(
I_{\mathrm{exist}},
I_{\mathrm{contain}},
\hat{\mathcal{E}},
\hat{\mathcal{R}},
n
\right),
\end{equation*}
where $I_{\mathrm{exist}}$ indicates whether the answer asserts any error, $I_{\mathrm{contain}}$ indicates whether it contains the annotated target error, $\hat{\mathcal{E}}$ and $\hat{\mathcal{R}}$ are the predicted evidence set and reasoning chain, and $n$ is the number of unrelated error claims. Let $\mathcal{E}^{*}$ and $\mathcal{R}^{*}$ denote the gold evidence set and reasoning chain.

The detection score requires the target error to be identified:
\begin{equation*}
S_{\mathrm{det}} =
I_{\mathrm{exist}} I_{\mathrm{contain}} .
\end{equation*}

The evidence location score measures overlap with the gold evidence while penalizing over-reporting:
\begin{equation*}
D_{\mathcal{E}} =
\frac{
2|\hat{\mathcal{E}}\cap\mathcal{E}^{*}|
+
\mathbf{1}\{|\hat{\mathcal{E}}|+|\mathcal{E}^{*}|=0\}
}{
\max(|\hat{\mathcal{E}}|+|\mathcal{E}^{*}|,1)
},
\end{equation*}
\begin{equation*}
S_{\mathrm{loc}} =
\max\left\{
0,\,
D_{\mathcal{E}}
-
0.8
\left(
\frac{
|\hat{\mathcal{E}}\setminus\mathcal{E}^{*}|
}{
\max(|\hat{\mathcal{E}}|,1)
}
\right)^2
\right\}.
\end{equation*}

For reasoning, ScholScan counts the matched prefix length between the predicted and gold reasoning chains:
\begin{equation*}
\hat{g} =
\operatorname{prefix\_match}
(\hat{\mathcal{R}}, \mathcal{R}^{*}),
\qquad
g_r = |\mathcal{R}^{*}|.
\end{equation*}
The reasoning score is then defined as
\begin{equation*}
S_{\mathrm{reason}} =
\mathbf{1}\{g_r=0\}
+
\mathbf{1}\{g_r>0\}
\left(
\frac{\hat{g}}{g_r}
\right)^2 .
\end{equation*}

To penalize unrelated error claims, it defines

\begin{equation*}
P_{\mathrm{unrel}}(n)
=
0.9^{\min(n,2)}
e^{-0.6[\max(n-2,0)]^{1.5}} .
\end{equation*}

The overall score combines target-error detection, evidence location, reasoning faithfulness, and the unrelated-error penalty:

\begin{equation*}
S(a)
=
S_{\mathrm{det}}
\cdot
\sqrt{
S_{\mathrm{location}}
\cdot
S_{\mathrm{reasoning}}
}
\cdot
P_{\mathrm{unrel}}(n).
\end{equation*}
\section{Reward Weight Analysis}
\label{app:appendix_e}

We further analyze the choice of reward weights through rollout statistics and post-hoc reweighting. For \textit{Reason} and \textit{Verify}, the evidence is specified, so we disable $R_{\text{alignment}}$ and use $(\omega_1,\omega_2,\omega_3)=(0.6,0,0.4)$, keeping reasoning completeness as the primary signal while retaining precision as a constraint against unsupported error claims. For \textit{Scan}, both issue and evidence cues are absent, so we use $(\omega_1,\omega_2,\omega_3)=(0.4,0.4,0.2)$, assigning symmetric primary weights to reasoning completeness and evidence alignment while using precision as an auxiliary constraint.

Table~\ref{tab:reward_variance_steps} reports the standard deviation of reward components over rollout steps. For \textit{Reason} and \textit{Verify}, $R_{\text{completeness}}$ consistently shows larger variance than $R_{\text{precision}}$, suggesting that it better differentiates reasoning quality across sampled trajectories. For \textit{Scan}, $R_{\text{completeness}}$ and $R_{\text{alignment}}$ have comparable variability, while $R_{\text{precision}}$ is more stable. This supports treating completeness and alignment as the main objectives in \textit{Scan}, with precision used to suppress excessive error reporting.

\begin{table*}[t]
\centering
\normalsize
\setlength{\tabcolsep}{5pt}
\caption{Standard deviation of reward components over rollout steps.}
\label{tab:reward_variance_steps}
\begin{tabular}{rccccccc}
\toprule
\multirow{2}{*}{\textbf{Step}} 
& \multicolumn{2}{c}{\textbf{\textit{Reason}}}
& \multicolumn{2}{c}{\textbf{\textit{Verify}}}
& \multicolumn{3}{c}{\textbf{\textit{Scan}}} \\
\cmidrule(lr){2-3}\cmidrule(lr){4-5}\cmidrule(lr){6-8}
& $\sigma(R_{\text{com}})$ & $\sigma(R_{\text{prec}})$
& $\sigma(R_{\text{com}})$ & $\sigma(R_{\text{prec}})$
& $\sigma(R_{\text{com}})$ & $\sigma(R_{\text{align}})$ & $\sigma(R_{\text{prec}})$ \\
\midrule
1  & 0.36 & 0.16 & 0.42 & 0.21 & 0.34 & 0.28 & 0.19 \\
5  & 0.24 & 0.14 & 0.34 & 0.20 & 0.28 & 0.23 & 0.17 \\
10 & 0.29 & 0.21 & 0.26 & 0.19 & 0.20 & 0.15 & 0.13 \\
15 & 0.32 & 0.17 & 0.32 & 0.18 & 0.28 & 0.24 & 0.14 \\
20 & 0.35 & 0.17 & 0.37 & 0.23 & 0.29 & 0.29 & 0.14 \\
25 & 0.28 & 0.13 & 0.32 & 0.18 & 0.30 & 0.25 & 0.16 \\
30 & 0.34 & 0.20 & 0.41 & 0.19 & 0.30 & 0.26 & 0.16 \\
\bottomrule
\end{tabular}
\end{table*}

We also perform post-hoc reweighting of $R_{\text{final}}$ without retraining, as shown in Table~\ref{tab:reward_reweighting_full}. The default setting corresponds to the weights used in the main experiments. The completeness-heavy setting increases the weight of $R_{\text{completeness}}$ to $(0.7,0,0.3)$ for \textit{Reason}/\textit{Verify} and $(0.5,0.3,0.2)$ for \textit{Scan}. We additionally test an equal-weight setting for \textit{Reason}/\textit{Verify}, using $(0.5,0,0.5)$. Across these variants, the main conclusions remain stable: the trained Qwen3-VL-8B model consistently improves over its base and SFT variants, and the relative difficulty of \textit{Scan} remains unchanged.

\begin{table*}[t]
\centering
\scriptsize
\setlength{\tabcolsep}{3pt}
\caption{Post-hoc reward reweighting results. Scores are scaled by 100. Default weights are $(0.6,0,0.4)$ for \textit{Reason}/\textit{Verify} and $(0.4,0.4,0.2)$ for \textit{Scan}. Completeness-heavy weights are $(0.7,0,0.3)$ for \textit{Reason}/\textit{Verify} and $(0.5,0.3,0.2)$ for \textit{Scan}. Equal R/V uses $(0.5,0,0.5)$ for \textit{Reason}/\textit{Verify}.}
\label{tab:reward_reweighting_full}
\resizebox{\textwidth}{!}{
\begin{tabular}{lcccccccc}
\toprule
\multirow{2}{*}{\textbf{Model}}
& \multicolumn{3}{c}{\textbf{Default}}
& \multicolumn{3}{c}{\textbf{Completeness-heavy}}
& \multicolumn{2}{c}{\textbf{Equal R/V}} \\
\cmidrule(lr){2-4}\cmidrule(lr){5-7}\cmidrule(lr){8-9}
& \textbf{Reason} & \textbf{Verify} & \textbf{Scan}
& \textbf{Reason} & \textbf{Verify} & \textbf{Scan}
& \textbf{Reason} & \textbf{Verify} \\
\midrule
Gemini 3 Pro 
& 60.0 & 54.5 & 24.3
& 59.8 & 54.0 & 24.8
& 60.2 & 55.1 \\
Qwen3-VL-235B-A22B Thinking 
& 51.8 & 46.8 & 17.4
& 52.5 & 45.4 & 17.8
& 51.0 & 48.3 \\
Qwen3-VL-235B-A22B Instruct 
& 45.9 & 14.2 & 3.4
& 46.1 & 13.9 & 3.5
& 45.7 & 14.6 \\
Qwen3-VL-8B Instruct 
& 29.5 & 5.7 & 2.0
& 28.0 & 5.3 & 2.1
& 31.1 & 6.2 \\
Qwen3-VL-8B SFT 
& 30.3 & 28.6 & 14.4
& 26.2 & 24.7 & 14.7
& 34.4 & 32.7 \\
Qwen3-VL-8B SFT+RL 
& 38.5 & 37.7 & 19.5
& 33.3 & 32.6 & 19.7
& 43.8 & 42.9 \\
\bottomrule
\end{tabular}
}
\end{table*}

These analyses do not replace full retraining under alternative rewards, but they show that the reported trends are not an artifact of one exact coefficient choice. The selected weights follow the cue structure of the tasks: evidence alignment is disabled when evidence is already specified, becomes central when evidence must be discovered, and precision remains a constraint against unsupported error reporting.
\section{Case Study}
\label{app:appendix_f}

\begin{figure*}[t]
  \centering
  \includegraphics[width=\linewidth]{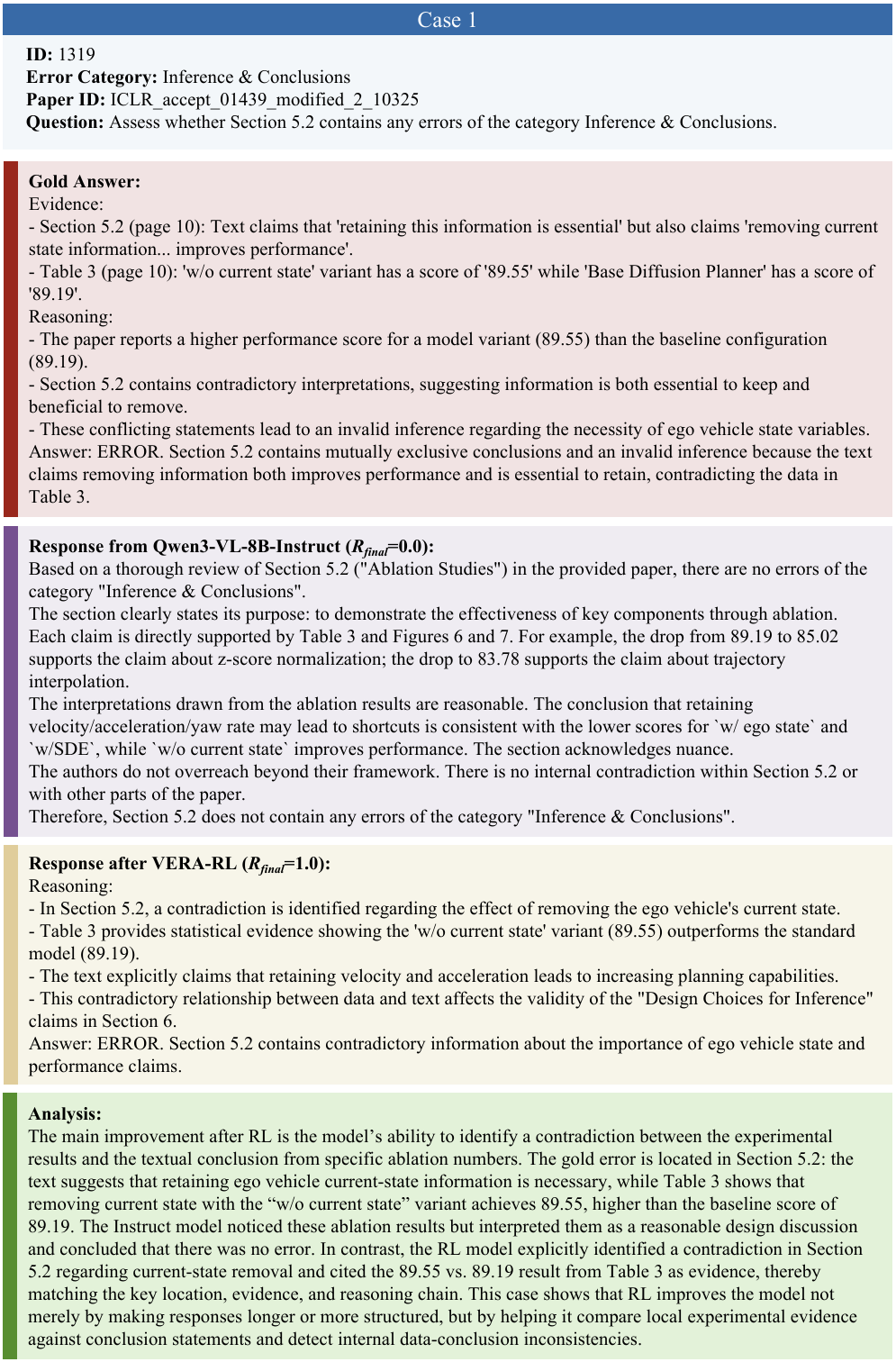}
  \caption{
    Case study on inference and conclusions.
  }
  \label{fig:Case1}
\end{figure*}

\begin{figure*}[t]
  \centering
  \includegraphics[width=\linewidth]{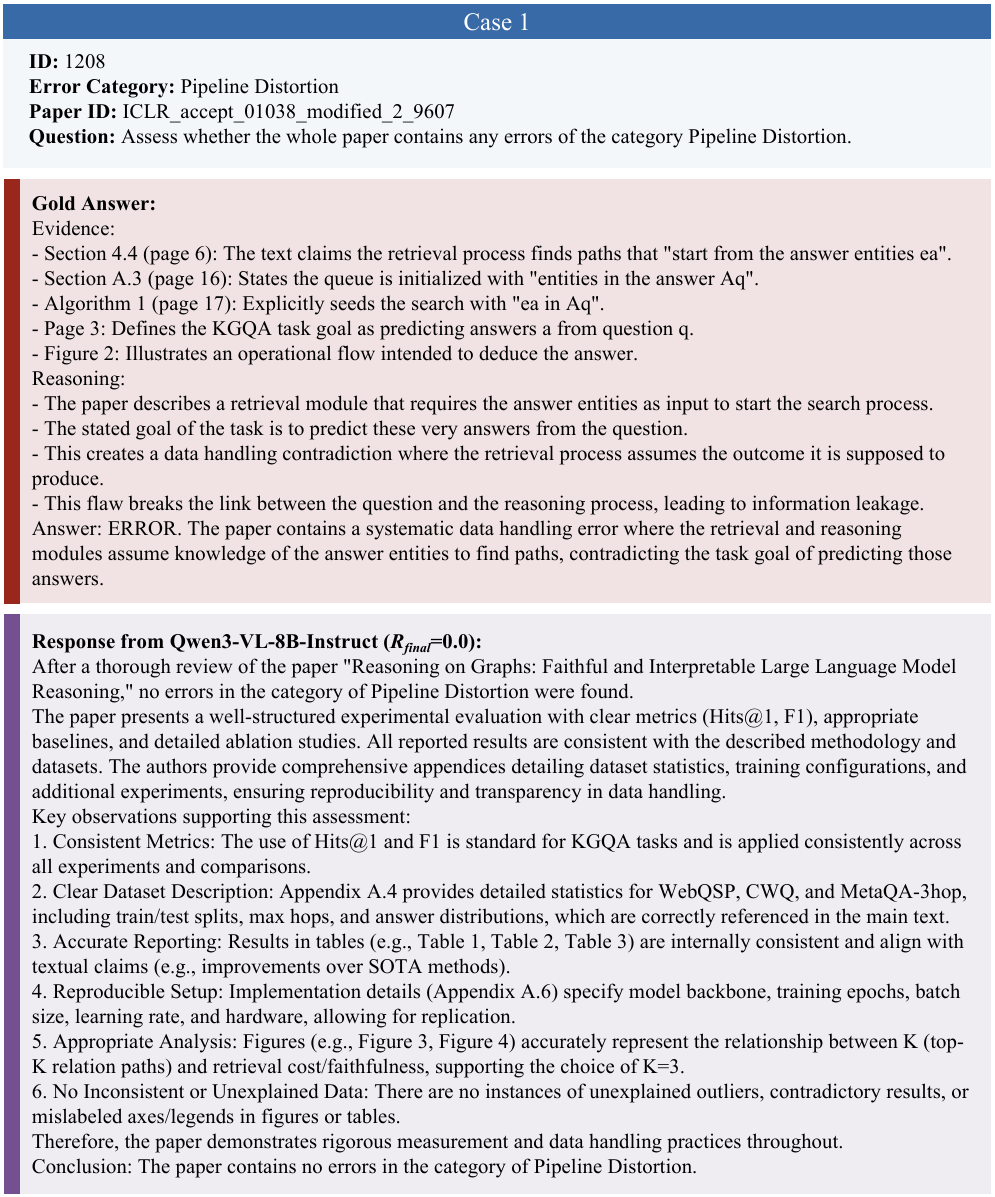}
  \caption{
    Case study on the first part of pipeline distortion.
  }
  \label{fig:Case2_1}
\end{figure*}

\begin{figure*}[t]
  \centering
  \includegraphics[width=\linewidth]{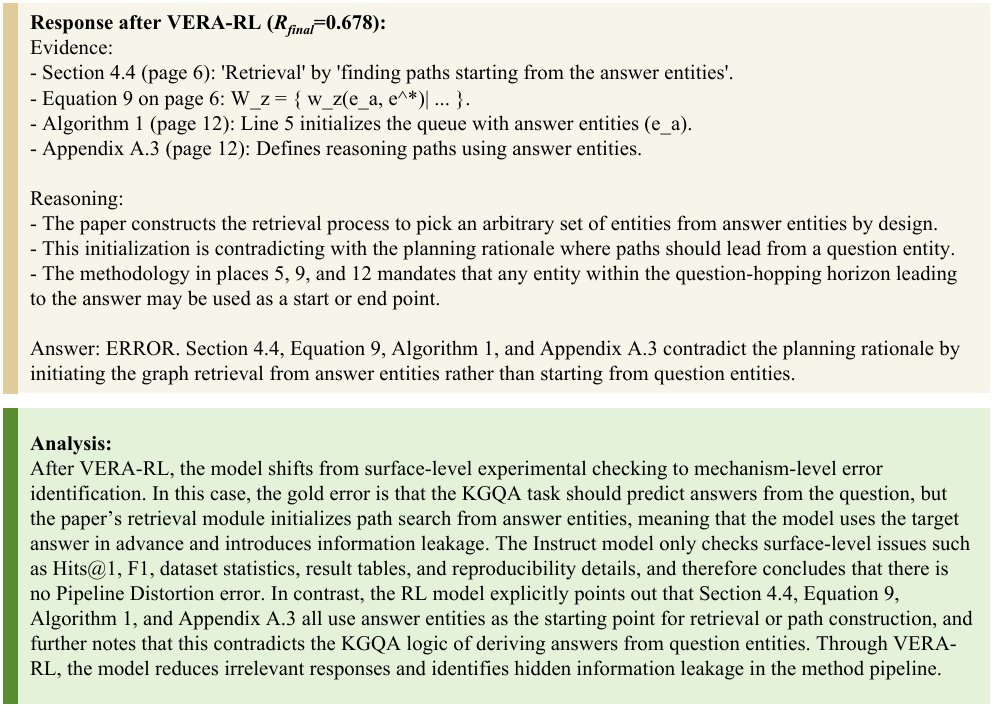}
  \caption{
    Case study on the second part of pipeline distortion.
  }
  \label{fig:Case2_2}
\end{figure*}

\begin{figure*}[t]
  \centering
  \includegraphics[width=\linewidth]{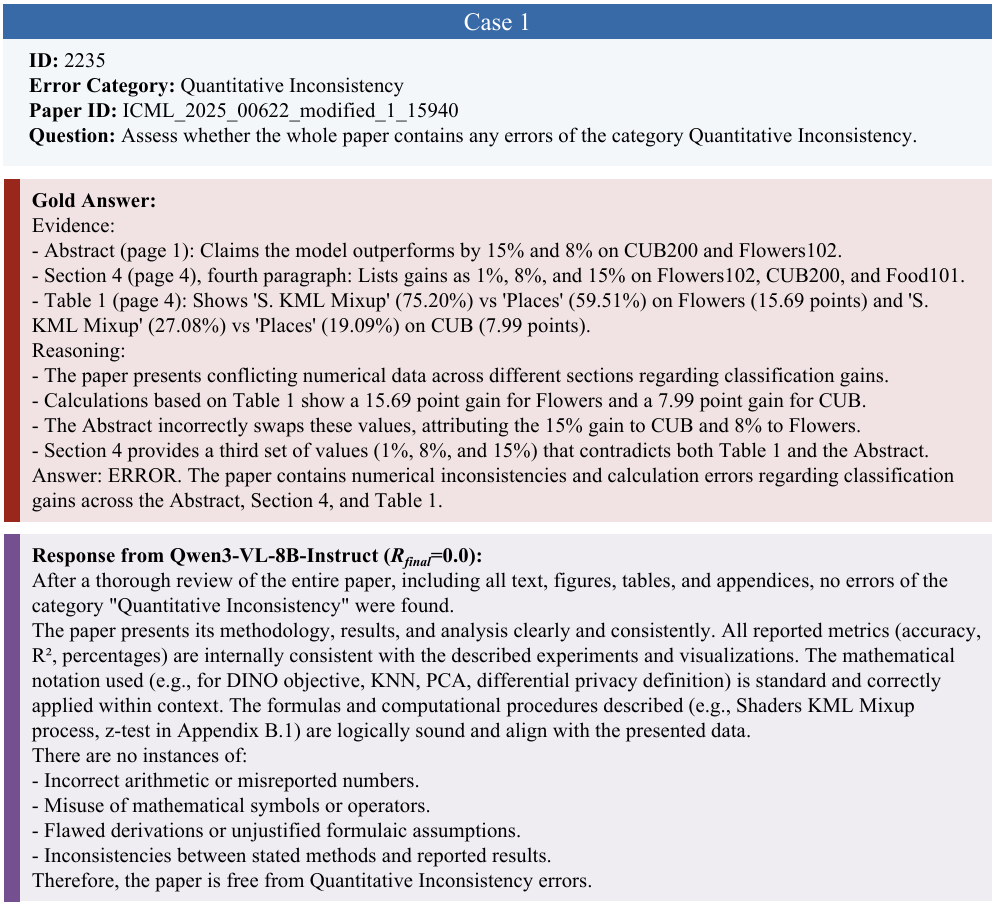}
  \caption{
    Case study on the first part of quantitative inconsistency.
  }
  \label{fig:Case3_1}
\end{figure*}

\begin{figure*}[t]
  \centering
  \includegraphics[width=\linewidth]{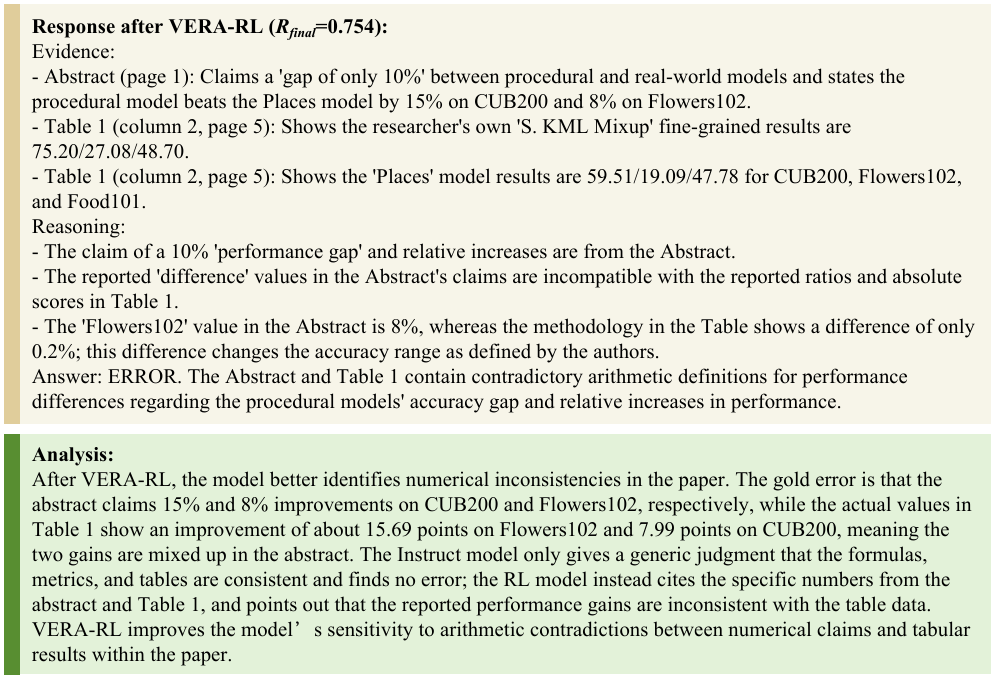}
  \caption{
    Case study on the second part of quantitative inconsistency.
  }
  \label{fig:Case3_2}
\end{figure*}

Figures~\ref{fig:Case1}, \ref{fig:Case2_1}, \ref{fig:Case2_2}, \ref{fig:Case3_1} and \ref{fig:Case3_2} provide qualitative examples showing how \textsc{VERA-RL} improves Qwen3-VL-8B-Instruct on Scan-style scientific error detection. Across the three cases, the original instruction-tuned model tends to give broad no-error judgments based on surface-level checks, such as whether metrics, tables, or experimental descriptions appear generally consistent. After VERA-RL, the model more often identifies the specific evidence needed for verification and connects it to the corresponding error. In the quantitative inconsistency case, it compares numerical claims in the abstract with table values and detects inconsistent performance gains. In the pipeline-distortion case, it traces the retrieval procedure across the main text, equation, algorithm, and appendix, identifying that the method starts from answer entities and therefore introduces information leakage. In the inference-and-conclusions case, it compares ablation results with the textual interpretation and detects a contradiction between the reported score and the claimed necessity of the current-state information. These examples illustrate that \textsc{VERA-RL} improves not only response format, but also evidence selection, cross-section consistency checking, and the ability to turn paper-level evidence into verifiable error judgments.

\section{Reliability Analysis}
\label{app:appendix_g}
\textbf{Cross-Evaluator Agreement.}
We independently rescore the step-30 RL rollouts using Qwen3-27B and Gemini 2.5 Flash. As shown in Table~\ref{tab:evaluator_agreement}, both evaluators show high correlation with the original scores across \textit{Reason}, \textit{Verify}, and \textit{Scan}, suggesting that the reward signals are not specific to a single evaluator.

\begin{table*}[t]
\centering
\caption{Correlation (\%) between reward scores extracted by alternative evaluators and those from Seed-1.6-Thinking on the step-30 RL rollouts.}
\begin{tabular}{lccc}
\toprule
Setup & Reason ($R_{\mathrm{final}}$) & Verify ($R_{\mathrm{final}}$) & Scan ($R_{\mathrm{final}}$) \\
\midrule
Qwen3-27B        & 78.3 & 88.4 & 88.7 \\
Gemini 2.5 Flash & 84.4 & 82.3 & 82.2 \\
\bottomrule
\end{tabular}
\label{tab:evaluator_agreement}
\end{table*}
\section{Reproducibility and Ethics Statement}
\label{app:appendix_h}
VERA-13K is constructed by the authors and does not reuse samples from previously released benchmarks. For papers from international conferences such as ICML, ICLR (including reviews), and NeurIPS, all content was crawled from the OpenReview platform. Papers from Nature Communications and PNAS were obtained entirely from open-access sources, ensuring that no privacy, ethical, or conflict-of-interest concerns arise.

The code and data are publicly available at
\url{https://github.com/Staudinger0325/VERA-RL}.
\end{document}